\RequirePackage[dvipsnames]{xcolor}
\PassOptionsToPackage{dvipsnames}{xcolor}

\documentclass{article}
\usepackage{amssymb}

\usepackage[final]{corl_2025} %

\usepackage{tikz}
\usepackage{amsmath}
\usepackage{algorithm}
\usepackage{algorithmic}
\usepackage{booktabs}
\usepackage{multirow}
\usepackage{subcaption}
\usepackage{changepage}
\usepackage{wrapfig}
\usepackage{pifont}
\usepackage{bm}

\newcommand{\cmark}{\textcolor{LimeGreen}{\ding{51}}}
\newcommand{\xmark}{\textcolor{red}{\ding{55}}}

\title{Mastering Multi-Drone Volleyball through Hierarchical Co-Self-Play Reinforcement Learning}

\author{
  Ruize Zhang, Sirui Xiang, Zelai Xu, Feng Gao, Shilong Ji, Wenhao Tang,  \\
  \textbf{Wenbo Ding, Chao Yu$^{\dagger}$, Yu Wang$^{\dagger}$}\\
  Tsinghua University\\
  \texttt{jimmyzhangruize@gmail.com, \{yuchao,yu-wang\}@mail.tsinghua.edu.cn}
}

\begin{document}
\maketitle

\let\thefootnote\relax
\footnotetext{$\dagger$ Corresponding authors.}

\begin{abstract}
In this paper, we tackle the problem of learning to play 3v3 multi-drone volleyball, a new embodied competitive task that requires both high-level strategic coordination and low-level agile control. The task is turn-based, multi-agent, and physically grounded, posing significant challenges due to its long-horizon dependencies, tight inter-agent coupling, and the underactuated dynamics of quadrotors. To address this, we propose Hierarchical Co-Self-Play (HCSP), a hierarchical reinforcement learning framework that separates centralized high-level strategic decision-making from decentralized low-level motion control. We design a three-stage population-based training pipeline to enable both strategy and skill to emerge from scratch without expert demonstrations: (I) training diverse low-level skills, (II) learning high-level strategy via self-play with fixed low-level controllers, and (III) joint fine-tuning through co-self-play. Experiments show that HCSP achieves superior performance, outperforming non-hierarchical self-play and rule-based hierarchical baselines with an average 82.9\% win rate and a 71.5\% win rate against the two-stage variant. Moreover, co-self-play leads to emergent team behaviors such as role switching and coordinated formations, demonstrating the effectiveness of our hierarchical design and training scheme.
The project page is at https://hi-co-self-play.github.io.

\end{abstract}


\section{Introduction}
Competitive tasks have long served as benchmarks for progress in artificial intelligence. Landmark results have been achieved in domains such as Go~\citep{silver2016mastering}, poker~\citep{moravvcik2017deepstack}, and real-time strategy games~\citep{vinyals2019grandmaster}, where agents learn to plan, adapt, and compete under structured rules. As research moves from virtual environments to the physical world, robot sports--structured, rule-based competitions involving physical agents--have emerged as a promising frontier for embodied intelligence. Examples include robot soccer~\citep{kitano1997robocup,liu2022motor}, table tennis~\citep{d2024achieving, ma2025mastering}, and multi-drone pursuit-evasion~\citep{chen2024multi}, which combine high-level strategy with low-level motion control in physically grounded settings.

In this paper, we tackle a new embodied competitive task proposed by the VolleyBots testbed~\citep{xu2025volleybots}: 3v3 multi-drone volleyball. This task exemplifies the structure of a robot sport--well-defined objectives, explicit rules, and head-to-head competition--while presenting a set of unique and underexplored challenges. Each team must coordinate three quadrotors to rally a ball over a net, switching roles dynamically between offense and defense in a turn-based fashion. The environment is highly dynamic and demands precise timing, agile 3D maneuvering, and strategic team-level behavior. The turn-based nature of ball exchange introduces long-horizon temporal dependencies; the multi-agent setting requires tightly coupled tactics; and the underactuated dynamics of quadrotors call for fine-grained, reactive motor skills. Compounding these challenges, no expert demonstrations are available, ruling out imitation learning~\citep{osa2018algorithmic} and requiring agents to discover both motion primitives and team strategies from scratch.

To address these challenges, we propose Hierarchical Co-Self-Play (HCSP), a hierarchical reinforcement learning (HRL) framework that separates strategic reasoning from motor control. The policy is decomposed into a high-level strategy responsible for team tactics, and multiple low-level skills responsible for drone-specific motion control. The high-level strategy is a centralized multi-layer perceptron (MLP) with three output heads, each producing commands for one drone, enabling synchronized decision-making across the team. The low-level policies are independent across drones and operate at a higher frequency to ensure responsive, fine-grained control. To reflect the turn-based structure of the game, the high-level strategy is only activated upon discrete game events--when the ball is struck or crosses the net--while low-level control runs continuously between these moments.

Training such a system end-to-end is challenging due to the tight coupling between high-level strategy and low-level control. To address this, we propose a three-stage training pipeline that enables both skills and strategies to emerge from scratch, without relying on any expert supervision. In Stage I, we train diverse low-level motion skills using reward shaping and policy chaining, an iterative training method that ensures smooth transitions between temporally adjacent skills. In Stage II, we freeze the low-level controllers and train the high-level strategy via population-based self-play and sample reallocation, allowing the team to learn fundamental cooperative and adversarial behaviors. In Stage III, we jointly fine-tune both high- and low-level policies through co-self-play, enabling mutual adaptation and further performance gains.

Our experiments show that the policy trained with HCSP significantly outperforms both standard self-play and rule-based hierarchical baselines. Beyond overall performance, we conduct a detailed analysis of the three-stage training pipeline. In particular, the final co-self-play stage (Stage III) leads to the emergence of sophisticated team behaviors, such as specialized formations and adaptive role switching between passing and attacking, which are absent in earlier stages. This results in a 71.5\% win rate over Stage II, validating the benefits of end-to-end hierarchical adaptation. Ablation studies further confirm the contribution of HCSP’s core components, including policy chaining in Stage I and sample reallocation in Stage II, each playing a critical role in enabling the emergence of effective and coordinated multi-agent behavior.

Our main contributions are as follows:
\begin{itemize}
\item We conduct the first systematic study of the 3v3 multi-drone volleyball task, which combines strategic team play, precise motion control, and high-speed aerial agility in a turn-based, multi-agent physical environment.

\item We propose Hierarchical Co-Self-Play (HCSP), a hierarchical reinforcement learning method that integrates a centralized high-level strategy for team tactics and decentralized low-level skills for agile control.

\item We introduce a three-stage population-based training pipeline that enables the emergence of both motion skills and tactical strategies without expert demonstrations. The final co-self-play stage fosters emergent behaviors such as role switching and formation specialization.

\item In extensive experiments, HCSP outperforms both non-hierarchical self-play and rule-based hierarchical baselines, achieving an 82.9\% average win rate against the baselines and a 71.5\% win rate against the two-stage variant.
\end{itemize}

\begin{figure}[t]
    \centering
    \includegraphics[width=\linewidth]{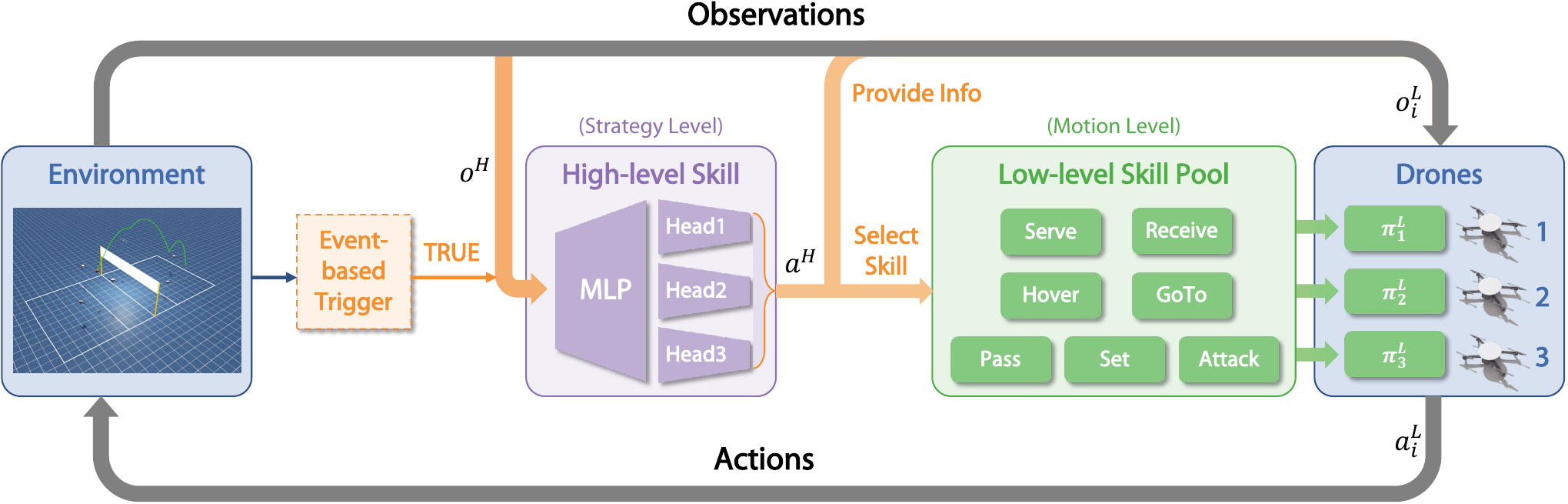}
    \caption{
    HCSP architecture: an event‐driven high‐level strategy handles strategic decisions, while multiple low‐level skills manage continuous control. Each drone \(i\) runs its low‐level skill at 50Hz, taking local observation \(o_i^L\) and high‐level tactical parameters to produce continuous action \(a_i^L\). The high‐level strategy activates only on discrete events (racket strike or ball crossing the net), observes \(o^H\), and outputs \(a^H\) to choose each drone’s skill and supply tactical information. It is implemented as a shared MLP with three output heads, one per drone.
    }
    \label{fig:system}
    \vspace{-4mm}
\end{figure}

\section{Preliminaries}
\subsection{3v3 Multi-Drone Volleyball Task}
VolleyBots~\citep{xu2025volleybots} introduces a drone volleyball testbed built on the OmniDrones~\citep{xu2024omnidrones} simulator, leveraging NVIDIA Isaac Sim's~\citep{makoviychuk2021isaac, mittal2023orbit} high-fidelity physics engine for efficient training. Our work focuses on VolleyBots's 3v3 competitive task, which slightly simplifies real-world volleyball rules for drone implementation.
Two teams of three drones each compete on a scaled-down 6m × 12m court while maintaining the official net height of 2.43m. Each team must return the ball across the net within three consecutive hits using coordinated actions. A team loses points for failing to return the ball, hitting out-of-bounds, violating turn-based interaction rules, or having drones collide with the ground. Each episode lasts a maximum of 15 seconds, starting with a randomized serving team. 
The episode ends when either one team scores or the time limit is reached.
The simulation employs Iris quadrotors~\citep{furrer2016rotors} equipped with horizontal 0.2m-radius rackets with 0.8 restitution coefficient. The drones are controlled via Per-Rotor Thrust (PRT) inputs at 50 Hz and receive state-based observations. The volleyball itself is modeled as a 5g sphere with 0.1m radius and 0.8 restitution coefficient to simulate realistic collision dynamics. See Appendix~\ref{appendix:1_task} for more details.

\subsection{Markov Game and MARL}

We model the 3v3 multi-drone volleyball task as a Partially Observable Markov Game (POMG)~\citep{shapley1953stochastic, littman1994markov}, defined by the tuple $\langle\mathcal{N},\mathcal{S},\{\mathcal{A}_i\}_{i\in\mathcal{N}},\{\mathcal{O}_i\}_{i\in\mathcal{N}}, P,\{R_i\}_{i\in\mathcal{N},}\gamma\rangle$.
$\mathcal{N}=\{1,\cdots,n\}$ denotes the set of agents. $\mathcal{S}$ represents the global state space. Each agent $i$ operates within an action space $\mathcal{A}_i$ and receives observations from an observation space $\mathcal{O}_i$. The state transition probability function $P: \mathcal{S} \times \{\mathcal{A}_i\}_{i\in\mathcal{N}} \times \mathcal{S}\rightarrow[0,1]$ defines the probability of transitioning from one state to another given the joint actions of all agents.  
$R_i:\mathcal{S}\times \mathcal{A}_i\rightarrow\mathbb{R}$ is the reward function of agent $i$, and $\gamma\in[0,1]$ is the discount factor determining the importance of future rewards. 
The 3v3 multi-drone volleyball task is a symmetric game. Both teams share the same continuous observation and action space and payoff structures (win or lose).
Multi-agent reinforcement learning (MARL) provides a natural framework for addressing Markov Games. In MARL, at each time step $t$, every agent $i$ receives an observation $o_{i,t}$ from the environment and selects an action $a_{i,t}$ according to its policy $\pi_i\in\Pi_i:\mathcal{O}_i\times\mathcal{A}_i\rightarrow[0,1]$. Given the joint action of all agents, the environment transitions from the current state $s_t$ to the subsequent state $s_{t+1}$ based on the transition function $P$ and returns an immediate reward $r_{i,t+1}$ to each agent $i$. Ultimately, the goal of agent $i$ is to learn a policy $\pi_i$ that maximize its expected discounted cumulative reward $\mathbb{E}_{\pi_{i}}[\sum_{t=0}^\infty \gamma^t r_{i,t+1}]$.

\section{Method}
The 3v3 multi-drone volleyball task presents dual challenges of precise motion control and coordinated team strategy. 
To address these requirements, we propose \textit{Hierarchical Co-Self-Play (HCSP)} (architecture in Fig.~\ref{fig:system}), a hierarchical reinforcement learning (HRL) method that decomposes the overall policy into an event-driven high-level strategy for team tactics and low-level skills for individual maneuvers.
HCSP's training pipeline comprises three stages:
(I) We train each low-level skill via RL using policy chaining, which addresses the challenge of executing temporally adjacent primitives smoothly by incrementally training each new skill in the context of previously learned ones, resulting in a robust pool of motion primitives.
(II) With the low-level skill pool frozen, we pretrain the high-level strategy through population-based team-level self-play, enabling it to acquire initial cooperative and competitive tactics.
(III) We utilize co-self-play to co-optimize both skill levels through team-level self-play, where the high-level strategy dynamically selects from continuously improving low-level skills while simultaneously refining their execution.

\subsection{Stage I: Low-Level Skill Acquisition}

Low-level skills focus on controlling individual drones to execute specific motion primitives without tactical reasoning. Each drone is regarded as a low-level agent. Agent $i$'s low-level skill observation $o_i^L (i\in\{1,\cdots,6\})$ consists solely of ball dynamics and the drone’s own state, together with high-level tactical parameters, but excludes information about other drones.
Following VolleyBots~\cite{xu2025volleybots}, which shows that Per-Rotor Thrust (PRT) slightly outperforms Collective Thrust and Body Rates (CTBR) in multi-drone volleyball by leveraging independent rotor control for greater agility, we adopt PRT as our low-level action space $a_i^L(i\in\{1,\cdots,6\})$. Low-level actions are executed at 50 Hz.
Additionally, we apply reward shaping to supply dense feedback and accelerate skill acquisition. For more details on the low-level skill design, see Appendix~\ref{appendix:2_stage1}.

Due to the challenge of ensuring smooth transitions between temporally adjacent low-level primitives, especially when control switches across drones or skill types, we employ policy chaining to iteratively train each low-level skill for seamless execution under the event-driven high-level strategy. For example, under policy chaining, the \textit{Attack} skill is trained following a teammate’s execution of the \textit{Set} skill, enabling coordinated multi-agent behaviors. This approach enhances intra-drone motion continuity and inter-drone cooperation.
Our experiments show that policy chaining significantly improves the reliability of sequential skill execution and accelerates learning. Detailed analysis is provided in Sec~\ref{sec:ablation_policy_chaining}. All low-level skills are optimized using Proximal Policy Optimization (PPO)~\cite{schulman2017proximal}, resulting in a diverse skill pool summarized in Table~\ref{tab:LLS}.

\begin{table}[t]
  \centering
  \caption{Description of seven low-level skills acquired in stage I and their corresponding tactical observations provided by the high-level strategy.}
  \vspace{1mm}
  \label{tab:LLS}
  \resizebox{\textwidth}{!}{
    \begin{tabular}{ccc}
        \toprule
        \textbf{Skill Name} & \textbf{Skill Description} & \textbf{Observation Provided by the High-Level Strategy} \\
        \midrule
        \textit{Serve} & Serve the ball toward the opponent’s court & Target ball position $(x,y,z)$ \\
        \textit{Receive} & Receive the serve ball and pass it to the setter & / \\
        \textit{Pass} & Pass the ball to the setter & Pass from court side $\{\text{left},\ \text{right}\}$ \\
        \textit{Set} & Place the ball optimally for the attacker & / \\
        \textit{Attack} & Strike the ball toward the opponent’s court & Attack direction \{left, right\} \\
        \textit{Hover} & Maintain at a fixed position and altitude & Preceding skill $\{\textit{Serve},\textit{Receive},\textit{Pass},\textit{Set},\textit{Attack}\}$ \\
        \textit{GoTo} & Move the drone to a specified location & Target drone position $(x,y,z)$ \\
        \bottomrule
      \end{tabular}
    }
    \vspace{-4mm}
\end{table}

\subsection{Stage II: High-Level Strategy Pretraining}

\begin{figure}[ht]
\vspace{-2mm}
  \centering
  \begin{subfigure}[b]{0.42\linewidth}
    \centering
    \includegraphics[width=\linewidth]{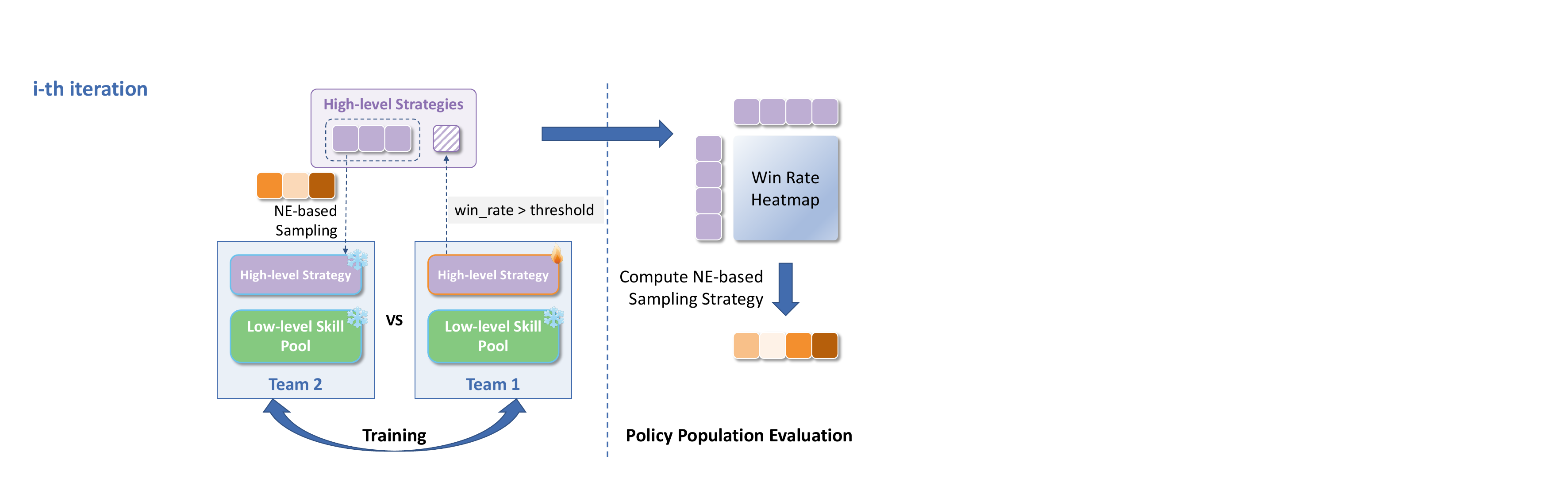}
    \caption{Full training schedule for Stage II.}
    \label{fig:stage2}
  \end{subfigure}%
  \hfill
  \begin{subfigure}[b]{0.55\linewidth}
    \centering
    \includegraphics[width=\linewidth]{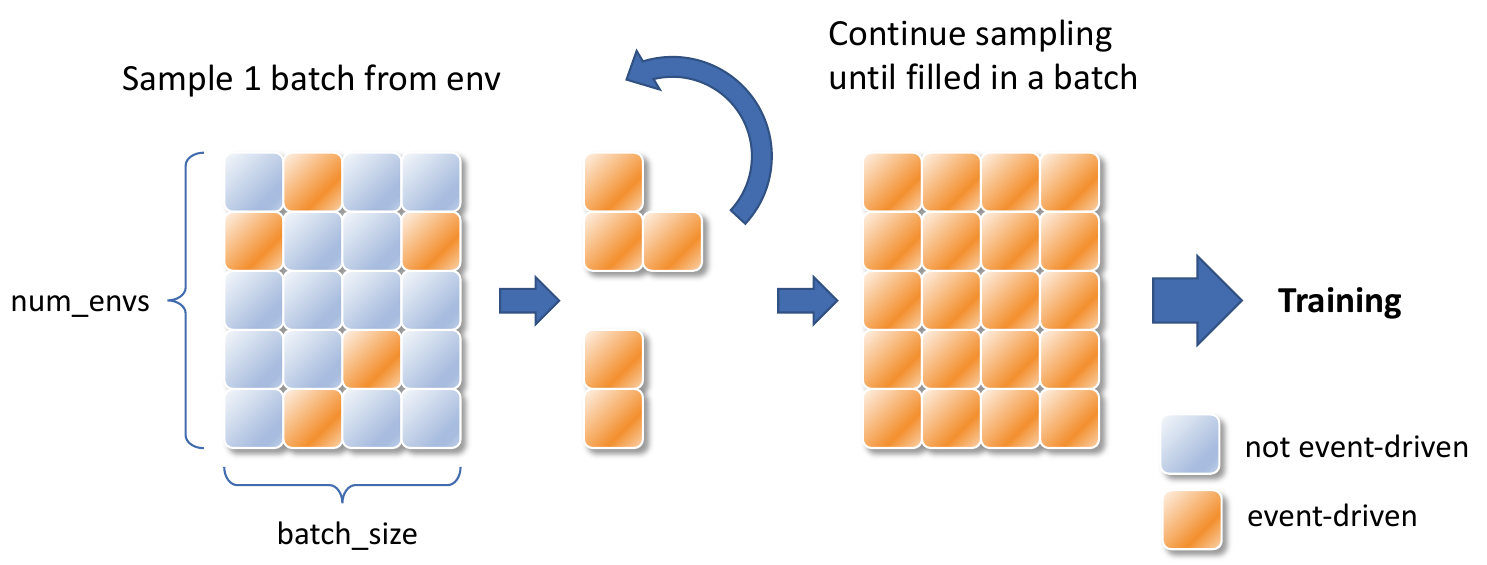}
    \caption{Sample reallocation for high-level strategy training.}
    \label{fig:reallocation}
  \end{subfigure}
  \caption{Illustrations of the high-level strategy pretraining stage (Stage II).}
  \label{fig:combined}
  \vspace{-2mm}
\end{figure}

High-level strategies enable team-level decision-making, with each team modeled as a centralized high-level agent. 
The observation $o_j^H$ ($j \in \{1, 2\}$) includes the states of all six drones (three teammates and three opponents), the ball dynamics, and other game-relevant features. 
At each decision point, the high-level strategy outputs an action $a_j^H$ that assigns a low-level skill and corresponding tactical parameters to each of its three drones. 
To support coordinated control, we implement the policy network as a shared multilayer perceptron (MLP) with three output heads, one per drone. 
Because the task is turn-based, high-level strategies are executed only at discrete events, such as when a racket hits the ball or when the ball crosses the net, indicating a possible change in team roles.
The reward for the high-level strategy $\pi_j^H$ at timestep $t$ is defined as:
\begin{equation}
    r_{j,t}^H = c_1 \times \mathrm{win\_or\_lose}_j + c_2 \times \mathrm{racket\_hit\_ball}_j,
\end{equation}
where $\mathrm{win\_or\_lose}_j = 1$ if team $j$ wins, $-1$ if it loses, and $0$ otherwise; $\mathrm{racket\_hit\_ball}_j = 1$ if any racket of team $j$ contacts the ball at timestep $t$, and $0$ otherwise.

In this stage, we freeze the low-level skills and pretrain the high-level strategy through team-level iterative self-play~\cite{lanctot2017unified} (Fig.~\ref{fig:stage2}). 
As the game is symmetric, both teams share the same high-level strategy population, and we train only one side against the other. 
At each iteration, a candidate high-level strategy competes against frozen opponents sampled from the population according to the Nash equilibrium (NE)-based distribution derived from the current win-rate matrix.
When the candidate’s win rate exceeds a threshold or a maximum step count is reached, it is added to the shared population.
We then perform pairwise evaluations of all population policies to produce an updated win-rate matrix, recompute the NE-based sampling distribution, and repeat the cycle.
Since the high-level strategy is event-driven, its action steps are temporally sparse, leading to high-variance gradients and unstable learning.  
To mitigate this issue, we apply \textit{sample reallocation} (Fig.~\ref{fig:reallocation}), which extracts transitions at event-triggered timesteps and attributes cumulative rewards over the intervals between events.
These event-triggered transitions are stored in a separate buffer for training.
The effectiveness of sample reallocation is analyzed in detail in Sec.~\ref{sec:ablation_sample_reallocation}.

\subsection{Stage III: Co-Self-Play}
\label{sec:stage3}

\begin{figure}[ht]
\vspace{-2mm}
    \centering
    \includegraphics[width=1.0\linewidth]{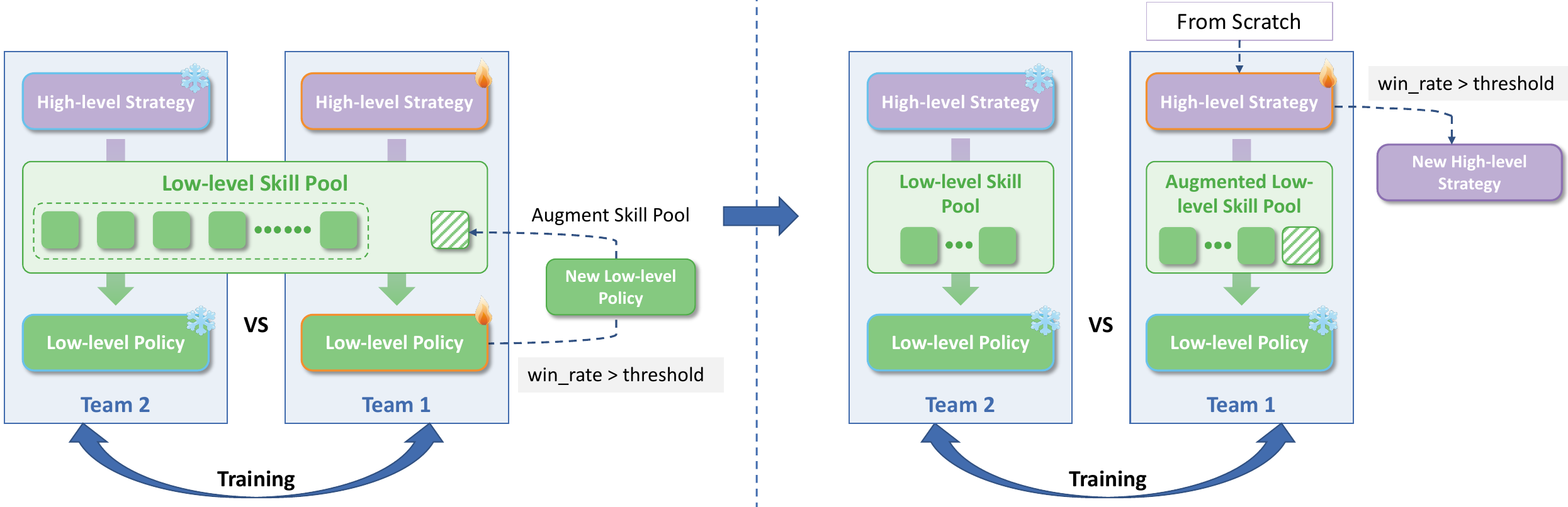}
    \caption{Illustration of the co-self-play stage (Stage III).}
    \label{fig:stage3}
    \vspace{-2mm}
\end{figure}

As high-level strategies can motivate low-level skill development and, in turn, improvements in low-level skills inform high-level decision-making, co-self-play is essential to iteratively optimize both layers.
In Stage III, we jointly refine both high- and low-level policies (Fig.~\ref{fig:stage3}). Similar to Stage II, we train only one team at a time against a frozen opponent, with the candidate high- and low-level skills learning together while the opponent’s policies remain frozen. Whenever the candidate’s win rate exceeds a threshold, the newly trained low-level policies are added to the low-level pool. 
As the low-level skill pool has grown, the high-level actor’s output layer needs to accommodate the expanded action set. We then freeze the augmented low-level pool and train a new high-level strategy from scratch against the former high-level strategy and the original low-level skill pool.

In Stage III, a key challenge in co-self-play lies in defining appropriate reward signals for low-level skills. 
A naive choice is to reuse the hand-crafted rewards from Stage I; however, this approach fails to account for two critical issues. 
First, the task setting fundamentally changes in a game context—for example, a ball that appears to be landing might still be intercepted by an opponent drone, making previous reward definitions inconsistent or even misleading. 
Second, Stage I rewards are task-specific and offer limited room for further evolution, thus constraining the emergence of advanced behaviors. 
Alternatively, manually redesigning reward functions for every skill under game dynamics would demand substantial expert effort and risks introducing bias. 
To overcome these limitations, we directly adopt the high-level reward as the learning signal for low-level skills during co-training. 
This approach is entirely free of manual design and promotes the emergence of skill adaptations driven by strategic outcomes, enabling the co-evolution of high-level tactics and low-level execution.
Because low-level actions execute at 50 Hz while high-level rewards occur only at discrete events, this reward signal is too sparse for effective low-level learning. To mitigate this, we add a KL-divergence penalty between the current low-level policy $\pi_i^L$ and its reference policy $\pi_{i,ref}^L$. 
Due to space limitation, we analyze the impact of this KL penalty in detail in Appendix~\ref{appendix:stage3/kl}.
The resulting co-self-play reward for low-level agent $i$ on team $j\in\{1,2\}$ at time $t$ is:
\begin{equation}
    r_{i,t}^L=r_{j,t}^H-c_3\times KL(\pi_i^L||\pi_{i,ref}^L).
\end{equation}

\section{Experiments}
We conduct a series of experiments to evaluate HCSP and elucidate its design. In particular, we seek to answer the following two questions:
(1) How does HCSP’s performance compare with baseline methods?
(2) What is the impact of Stage III co-self-play on overall policy effectiveness?
(3) Which HCSP components are the key elements to its success?

\noindent\textbf{Baselines.} We evaluate HCSP against five baselines following VolleyBots~\citep{xu2025volleybots}. Four of them are non-hierarchical game-theoretic self-play methods: Self-Play (SP)~\citep{samuel1959some}, Fictitious Self-Play (FSP)~\citep{heinrich2015fictitious}, and two variants of Policy-Space Response Oracles (PSRO)~\citep{lanctot2017unified} using either a uniform meta-solver ($\text{PSRO}_{\text{Uniform}}$) or a Nash meta-solver ($\text{PSRO}_{\text{Nash}}$). The fifth is a rule-based hierarchical policy included in VolleyBots, which we denote as “Bot”. See Appendix~\ref{appendix:5_baselines} for a detailed baseline description.

\noindent\textbf{Evaluation metrics.} Head-to-head win rate, evaluated over 500 trajectories per matchup, serves as our primary metric for comparing relative policy strength. However, it may fail to reflect absolute strength due to potential cyclic dominance in non-transitive games. As noted by \citet{liu2019emergent}, although Elo rating~\citep{elo1978rating} is commonly used for this purpose, it is sensitive to the policy population. To mitigate this, they propose the Nash-averaging evaluator, which computes a Nash equilibrium over randomly sampled ten policies. We adopt it as our secondary metric in Sec.~\ref{sec:quant_analysis_of_stage_three}. All evaluations are averaged over three seeds. Details of implementation can be found in Appendix~\ref{appendix:6_evaluation_metrics}.

\subsection{Game Results Against Baselines}

\begin{figure}[t]
\vspace{-4mm}
  \centering
  \begin{subfigure}[b]{0.33\linewidth}
    \centering
    \includegraphics[width=\linewidth]{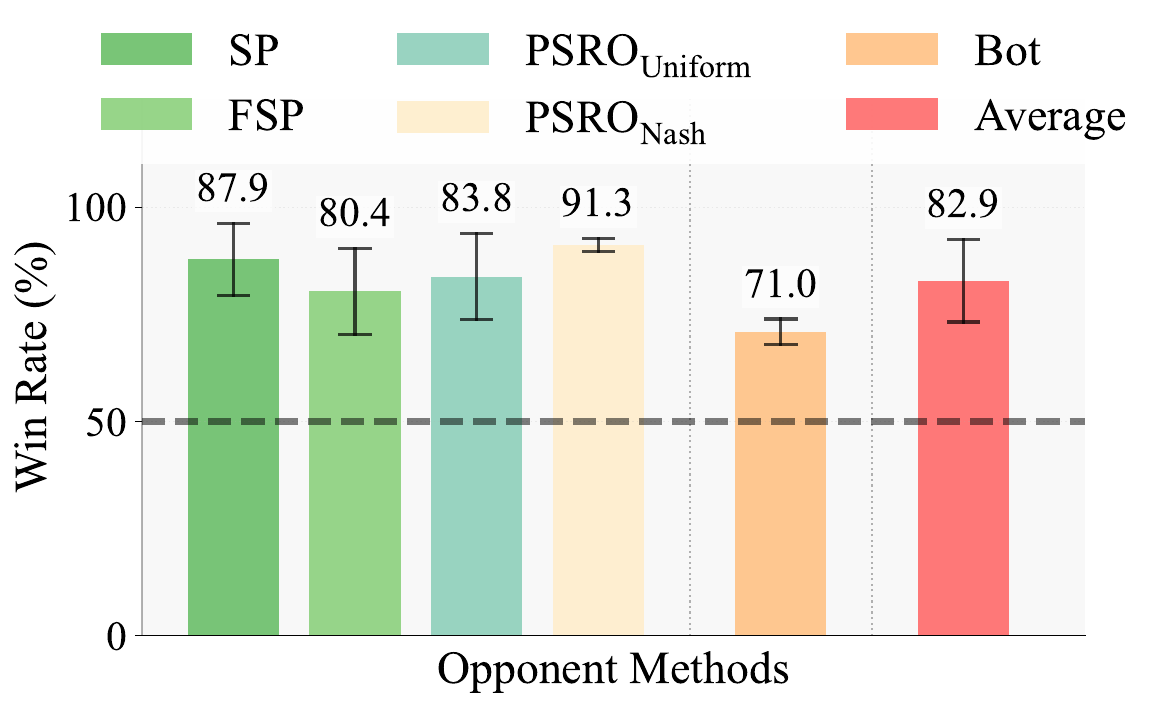}
    \caption{}
    \label{fig:baseline_sub}
  \end{subfigure}%
  \hfill
  \begin{subfigure}[b]{0.23\linewidth}
    \centering
    \includegraphics[width=\linewidth]{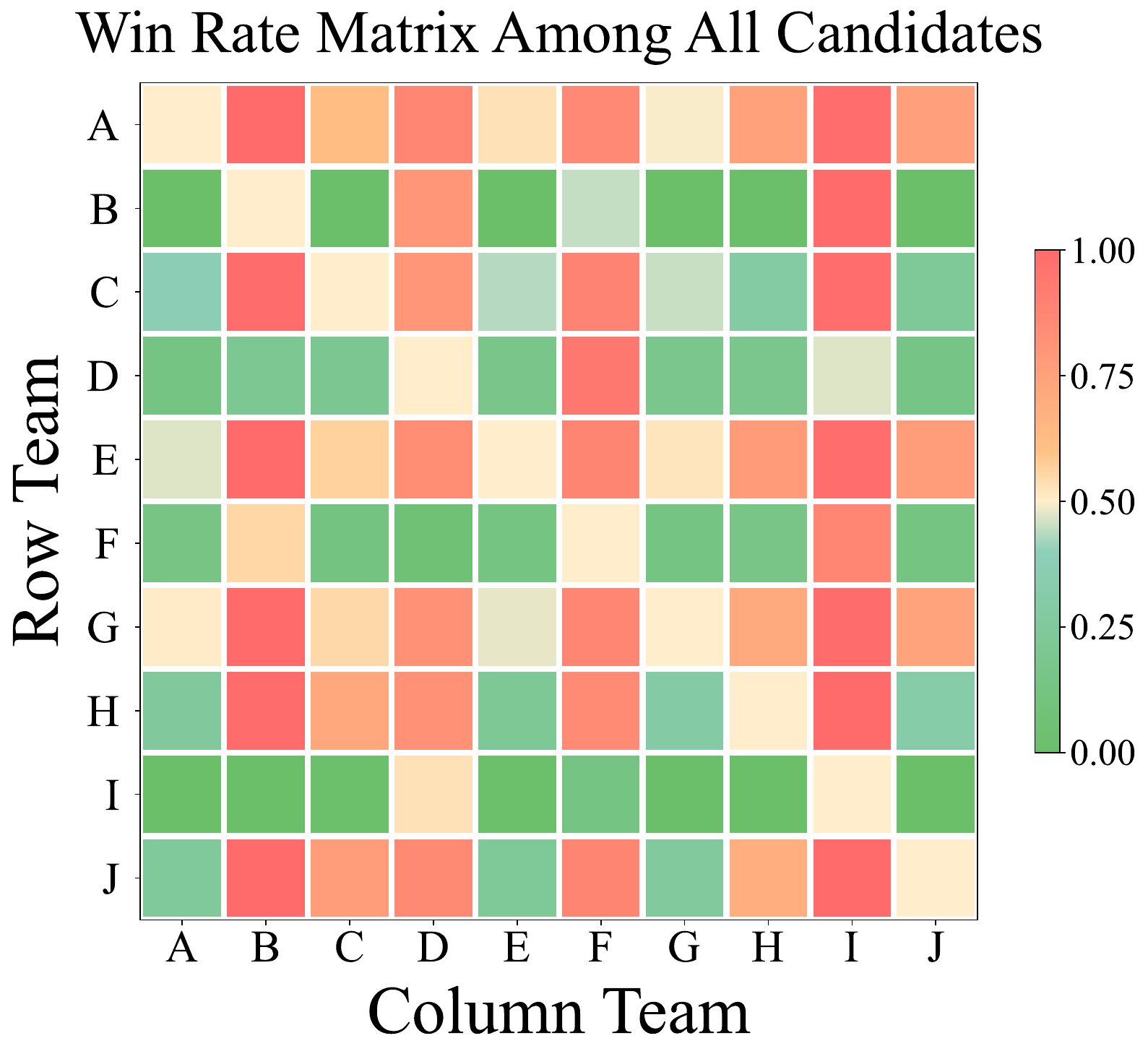}
    \caption{}
    \label{fig:nash_averaging_sub}
  \end{subfigure}
  \hfill
  \begin{subfigure}[b]{0.4\linewidth}
    \centering
    \includegraphics[width=\linewidth]{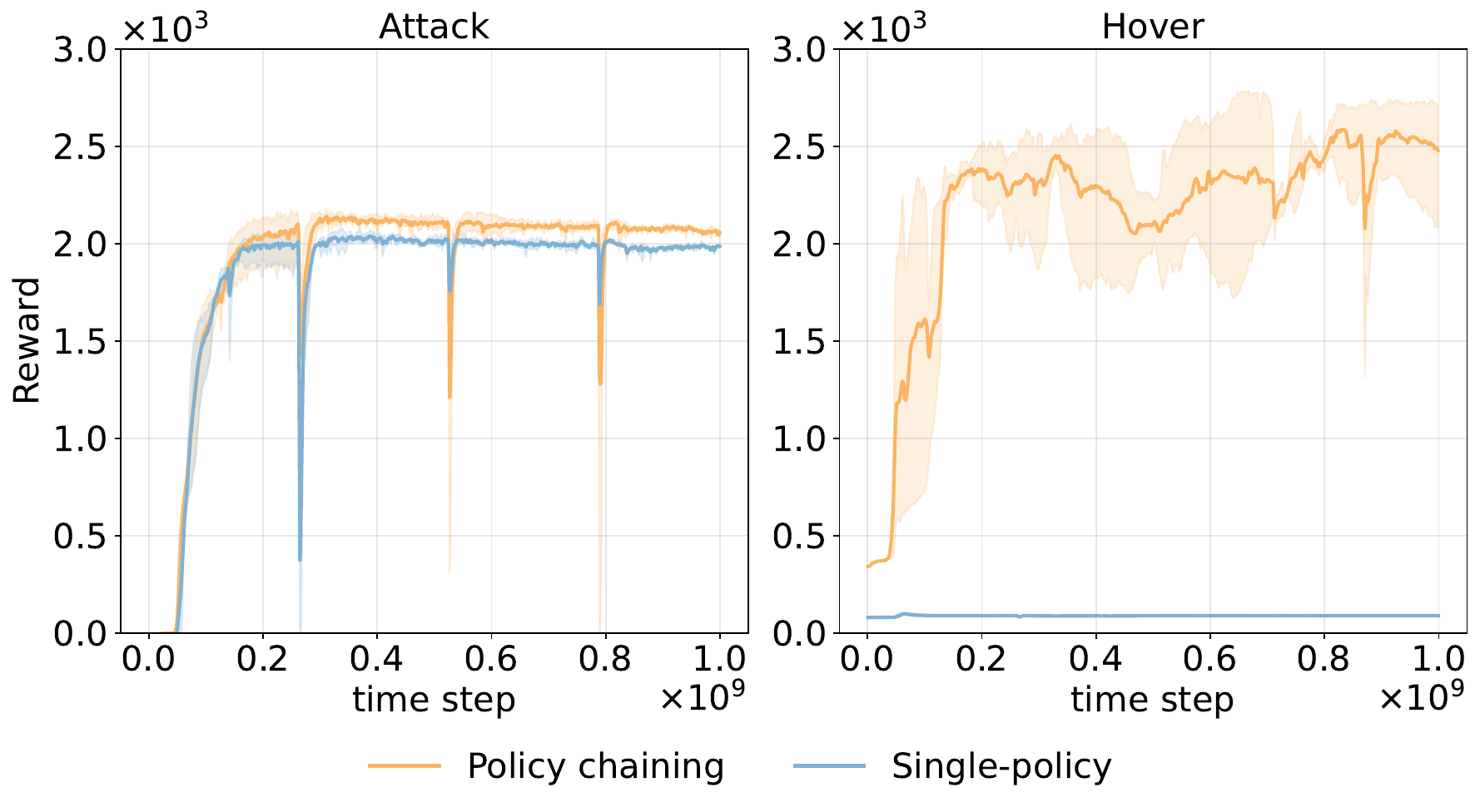}
    \caption{}
    \label{fig:att_hover_sub}
  \end{subfigure}
  \vspace{-2mm}
  \caption{Experiment results. (a) HCSP performance against baseline policies. (b) Cross-play win rate heatmap among ten randomly picked policies. (c) Ablation study on policy chaining in Stage I: policy chaining enables successful transition from Attack to Hover, whereas single-policy fails.}
  \label{fig:experiment}
  \vspace{-5mm}
\end{figure}

Fig.~\ref{fig:baseline_sub} shows the win rates when HCSP plays against each of the five baseline methods. HCSP achieves an average win rate of 82.9\% across all matchups, demonstrating its strong competitive advantage. Notably, it significantly outperforms all four non-hierarchical game-theoretic methods. Moreover, HCSP attains a 71\% win rate against the strongest baseline, the rule-based hierarchical Bot policy provided in VolleyBots~\cite{xu2025volleybots}. These results indicate that HCSP both outperforms traditional self-play strategies and competes strongly against the handcrafted hierarchical policy.

\subsection{Analysis of Stage III Co-Self-Play}


\subsubsection{Quantitative Analysis}
\label{sec:quant_analysis_of_stage_three}

To evaluate the effectiveness of Stage III training, we first conduct a head-to-head tournament between the Stage II and Stage III policies, as shown in the first two columns of Tab.~\ref{tab:stage2_vs_stage3}. The Stage III policy achieves a 71.5\% win rate over the Stage II policy, demonstrating clear improvement under the co-self-play framework.
However, this comparison only reflects relative performance between two policies. To assess absolute robustness, we construct a Nash-averaging evaluator following \citet{liu2019emergent}. Specifically, we randomly sample ten policies from all trained populations, including baselines and all HCSP stages, and compute their pairwise win rates (Fig.~\ref{fig:nash_averaging_sub}). We then apply Nash averaging to obtain a mixed evaluation policy, which assigns non-zero weights to three policies \{A (0.37), E (0.07), G (0.56)\} while assigning zero weight to all others.
Finally, we evaluate both the Stage II and Stage III policies against this Nash-averaged opponent. As shown in Tab.~\ref{tab:stage2_vs_stage3}, the Stage III policy improves its win rate from 31.4\% (Stage II) to 47.6\%, a 16.2\% absolute gain. Notably, the Stage III policy reaches approximately 50\% win rate against the Nash-averaged evaluator, indicating competitive performance near the Nash Equilibrium of the sampled strategy set.

\begin{table}[ht]
  \centering
  \vspace{-5mm}
  \caption{Win rates of Stage II policy and Stage III policy against different opponents.}
  \vspace{1mm}
  \label{tab:stage2_vs_stage3}
  \begin{tabular}{ccc|c}
    \toprule
    & Stage II policy  & Stage III policy  & Nash-averaged Policy \\
    \midrule
    Stage II policy & $50.0 \pm 0.0$ & $28.5\pm8.0$ & $31.4\pm9.3$ \\ 
    Stage III policy & $\mathbf{71.5 \pm 8.0}$ & $\mathbf{50.0\pm0.0}$ & $\mathbf{47.6\pm1.0}$ \\
    \bottomrule
  \end{tabular}
  \vspace{-4mm}
  
\end{table}

\subsubsection{Emergent Low-Level Skill}

\begin{figure}[t]
\vspace{-4mm}
  \centering
  \begin{subfigure}[b]{0.16\textwidth}
    \centering
    \includegraphics[width=\linewidth]{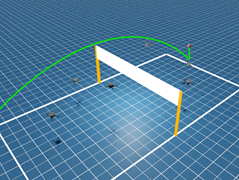}
    \caption{}
    \label{fig:2a}
  \end{subfigure}\hfill
  \begin{subfigure}[b]{0.16\textwidth}
    \centering
    \includegraphics[width=\linewidth]{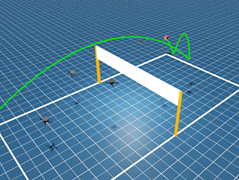}
    \caption{}
    \label{fig:2b}
  \end{subfigure}\hfill
  \begin{subfigure}[b]{0.16\textwidth}
    \centering
    \includegraphics[width=\linewidth]{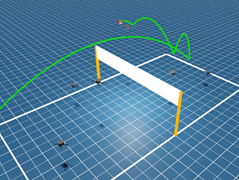}
    \caption{}
    \label{fig:2c}
  \end{subfigure}\hfill
  \begin{subfigure}[b]{0.16\textwidth}
    \centering
    \includegraphics[width=\linewidth]{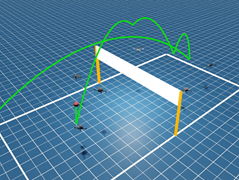}
    \caption{}
    \label{fig:2d}
  \end{subfigure}\hfill
  \begin{subfigure}[b]{0.16\textwidth}
    \centering
    \includegraphics[width=\linewidth]{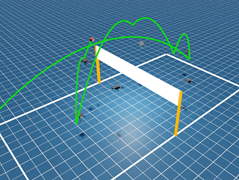}
    \caption{}
    \label{fig:2e}
  \end{subfigure}\hfill
  \begin{subfigure}[b]{0.16\textwidth}
    \centering
    \includegraphics[width=\linewidth]{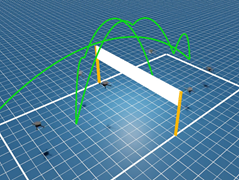}
    \caption{}
    \label{fig:2f}
  \end{subfigure}
  \vspace{-1mm}
  \caption{Sequence of six temporally sampled frames illustrating an emergent team behavior. (a) The opponent team \textit{passes} the ball to the setter. (b) The opponent team \textit{sets} the ball to the attacker. (c) The opponent team \textit{attacks} the ball towards our side. (d) Our team \textit{passes} the ball to the setter. (e) Our team performs a ``dump'' shot, sending the ball directly onto the opponent’s court. (f) The opponent team fails to return the ball, so our team scores the point.}
  \label{fig:six_horizontal}
  \vspace{-5mm}
\end{figure}

From a qualitative perspective, the experimental results further show that Stage III's co-self-play induces emergent low-level behaviors that were not designed in the initial stage. In Stage I, policy chaining yields three motion primitives, \textit{Pass}, \textit{Set}, and \textit{Attack}, with high accuracy, so each team always uses three consecutive contacts (\textit{Pass} to the setter drone, \textit{Set} to the attacker drone, \textit{Attack} over the net). During Stage II, with low-level skills frozen, the high-level strategy continues to learns this three-touch sequence. In contrast, after the Stage III, we observe a novel two-touch maneuver, akin to the real-world volleyball ``dump'' shot, in which the setter itself redirects the ball over the net instead of raising it for a teammate. 
This emergent ``dump'' skill requires no additional reward design and illustrates two key effects of co-self-play: low-level primitives can evolve beyond their initial definitions, and the high-level strategy adapts its skill allocation (for example, omitting \textit{Attack} and assigning \textit{Hover} to the would-be attacker drone) to leverage these new behaviors. Fig.~\ref{fig:six_horizontal} presents a case study in which one team uses the emergent ``dump'' skill to score a point.

\subsection{Ablation Study}

\subsubsection{Policy Chaining in Stage I}
\label{sec:ablation_policy_chaining}

In Stage I, we employ policy chaining, an iterative training method designed to ensure smooth transitions between temporally adjacent low-level skills. We compare policy chaining with a single-policy approach by training the ``\textit{Attacking} the ball to the opponent's side, followed by \textit{Hovering} to maintain stability'' behavior under the same reward settings. 
Through policy chaining, we first train an \textit{Attack} policy using \textit{Attack}-specific rewards, then train a subsequent \textit{Hover} policy with \textit{Hover}-specific rewards. In contrast, the single-policy approach trains one unified policy using the combined \textit{Attack} and \textit{Hover} rewards. 
In Fig.~\ref{fig:att_hover_sub}, they achieve similar performance in \textit{Attack}. However, only policy chaining successfully learns \textit{Hover} after \textit{Attack}, demonstrating that policy chaining reduces exploration difficulty by decomposing the complex behavior into sequential, specialized policies.

\subsubsection{High-level Strategy Sample Reallocation in Stage II}
\label{sec:ablation_sample_reallocation}

\begin{wrapfigure}{r}{0.53\textwidth}  %
  \vspace{-4mm}                        %
  \centering
  \begin{subfigure}[b]{0.47\linewidth}
    \centering
    \includegraphics[width=\linewidth]{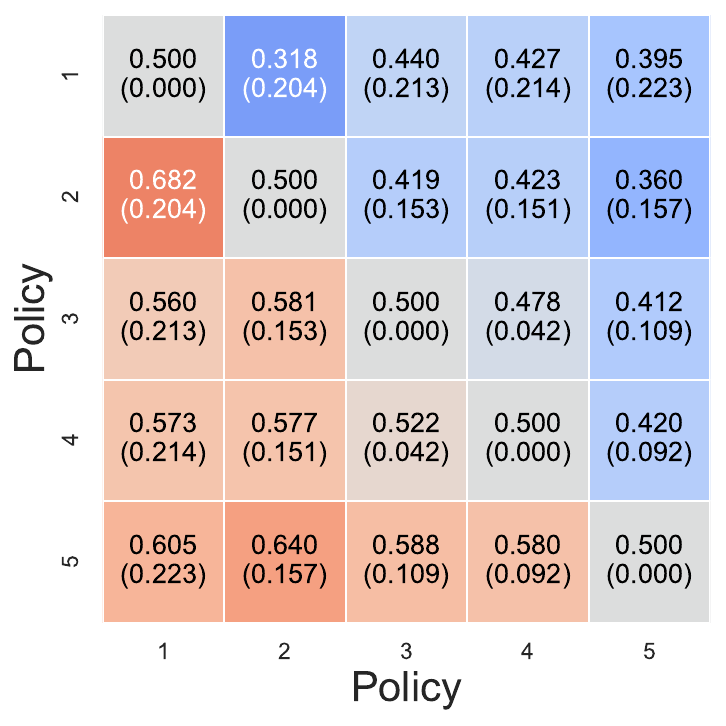}
    \caption{w/o sample reallocation.}
    \label{fig:stage_2_ablation_heatmap}
  \end{subfigure}
  \hfill
  \begin{subfigure}[b]{0.47\linewidth}
    \centering
    \includegraphics[width=\linewidth]{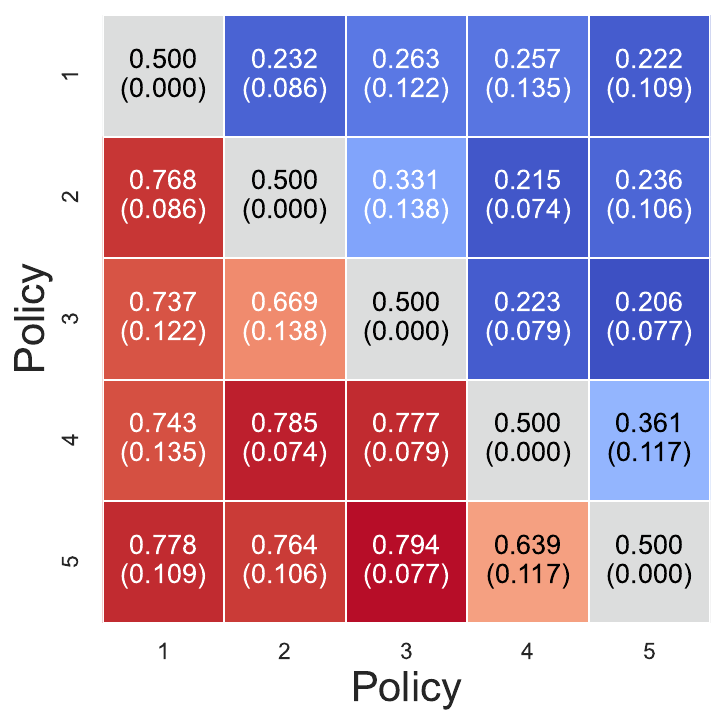}
    \caption{w/ sample reallocation.}
    \label{fig:stage_2_heatmap}
  \end{subfigure}
  \caption{Comparison of policy population evolution win-rate heatmaps in Stage II.}
  \vspace{-3mm}
\end{wrapfigure}

Because the high-level strategy is activated only at discrete time steps, standard minibatch sampling yields sparse effective samples and high variance. To address this, we introduce the sample reallocation procedure that reconstructs each training batch exclusively from event-driven steps (Fig.~\ref{fig:reallocation}). 
We compare standard minibatch sampling to our sample reallocation method by visualizing the high-level strategy’s evolution over five self-play iterations using win-rate heatmaps (Figs.~\ref{fig:stage_2_ablation_heatmap} and~\ref{fig:stage_2_heatmap}). In these heatmaps, off‑diagonal entries reflect each new policy’s win rate against its predecessors. Using sample reallocation, these off‑diagonal cells become strongly favorable much earlier, indicating faster skill progression, and exhibit lower win‐rate variance, indicating more stable training.
Also, we compare two policy populations in 500 head‑to‑head matches averaged over three random seeds, where each side samples a strategy from its population according to its NE–based distribution at the start of each match. The sample reallocation population wins $90.5\%\pm4.5\%$ of matches against the standard‑sampling population.

\section{Related Work}
The domain of robot sports has become a critical frontier for robotics research, demanding advances in real-time decision-making, dynamic control, and multi-agent collaboration. Initiatives such as RoboCup~\citep{kitano1997robocup} have driven progress in robot football, enabling humanoid and quadruped platforms to compete in full-scale matches~\citep{liu2019emergent,liu2022motor,haarnoja2024learning,xiong2024mqe}, while other systems have demonstrated human-level performance in table tennis~\citep{d2024achieving}, drone racing~\citep{kaufmann2023champion}, and beyond. Yet these benchmarks tackle only isolated aspects—team tactics, split-second precision, or high-speed maneuvering—whereas the 3v3 multi-drone volleyball task combines all of these under a turn-based game structure that existing methods cannot fully address. Hierarchical control architectures offer a promising solution by managing complexity across levels of abstraction: traditional behavior-based approaches~\citep{arkin1998behavior} and recent learning-based frameworks such as HiREPS~\citep{daniel2016hierarchical} and SayCan~\citep{ahn2022can} show how decomposing tasks into low-level skills and high-level planners improves adaptability and sample efficiency. 
Building on these insights, our work introduces an event-driven hierarchical policy that decomposes the volleyball task into a centralized high-level strategy for team tactics and multiple drone-specific low-level skills for motion control, enabling both rapid aerial agility and coordinated tactics.

\section{Conclusion}
To solve the 3v3 multi-drone volleyball task, we propose Hierarchical Co-Self-Play (HCSP), a novel hierarchical reinforcement learning framework. 
By decomposing policy into an event-driven high-level strategy for sparse strategic decisions and a suite of low-level skills for dense motion control, HCSP separates the challenges of team tactics, fine-grained control, and high-speed aerial agility. Our three-stage, demonstration-free training pipeline enables tactical strategies and agile controllers to emerge from scratch. Empirical results demonstrate that HCSP significantly outperforms baselines, and the co-self-play stage even triggers emergent behaviors without additional reward design.

\clearpage

\subsubsection*{Limitations}
Our approach is currently limited to high-fidelity simulation, which cannot fully capture real-world sensor noise, or hardware latency. Consequently, HCSP-trained policies may encounter a sim-to-real gap and are unlikely to transfer zero-shot to physical drones. Nonetheless, we report preliminary sim-to-real results in Appendix~\ref{appendix:7_preliminary_real_world_experiments}, which provide promising initial validation in real-world experiments.
Furthermore, HCSP assumes access to full state information (drone poses, velocities, and ball dynamics) rather than raw sensor or vision inputs, restricting its applicability in settings where only onboard cameras are available.

\acknowledgments{We sincerely thank Jiayu Chen, Chuqi Wang, and Yinuo Chen for their insightful discussions and experimental assistance throughout the course of this work. We are also grateful to Sicheng He for his initial involvement in the project. Their support and thoughtful input have contributed to clarifying our ideas and enhancing the overall quality of the paper.

This research was supported by National Natural Science Foundation of China (No.62406159, 62325405), Postdoctoral Fellowship Program of CPSF under Grant Number (GZC20240830, 2024M761676), China Postdoctoral Science Special Foundation 2024T170496.
}

\bibliography{reference}

\clearpage

\appendix

\section{Details of the 3v3 Muiti-Drone Volleyball Task}
\label{appendix:1_task}
\begin{figure}[h]
  \centering
  \begin{subfigure}[b]{0.5\linewidth}
    \centering
    \includegraphics[width=\linewidth]{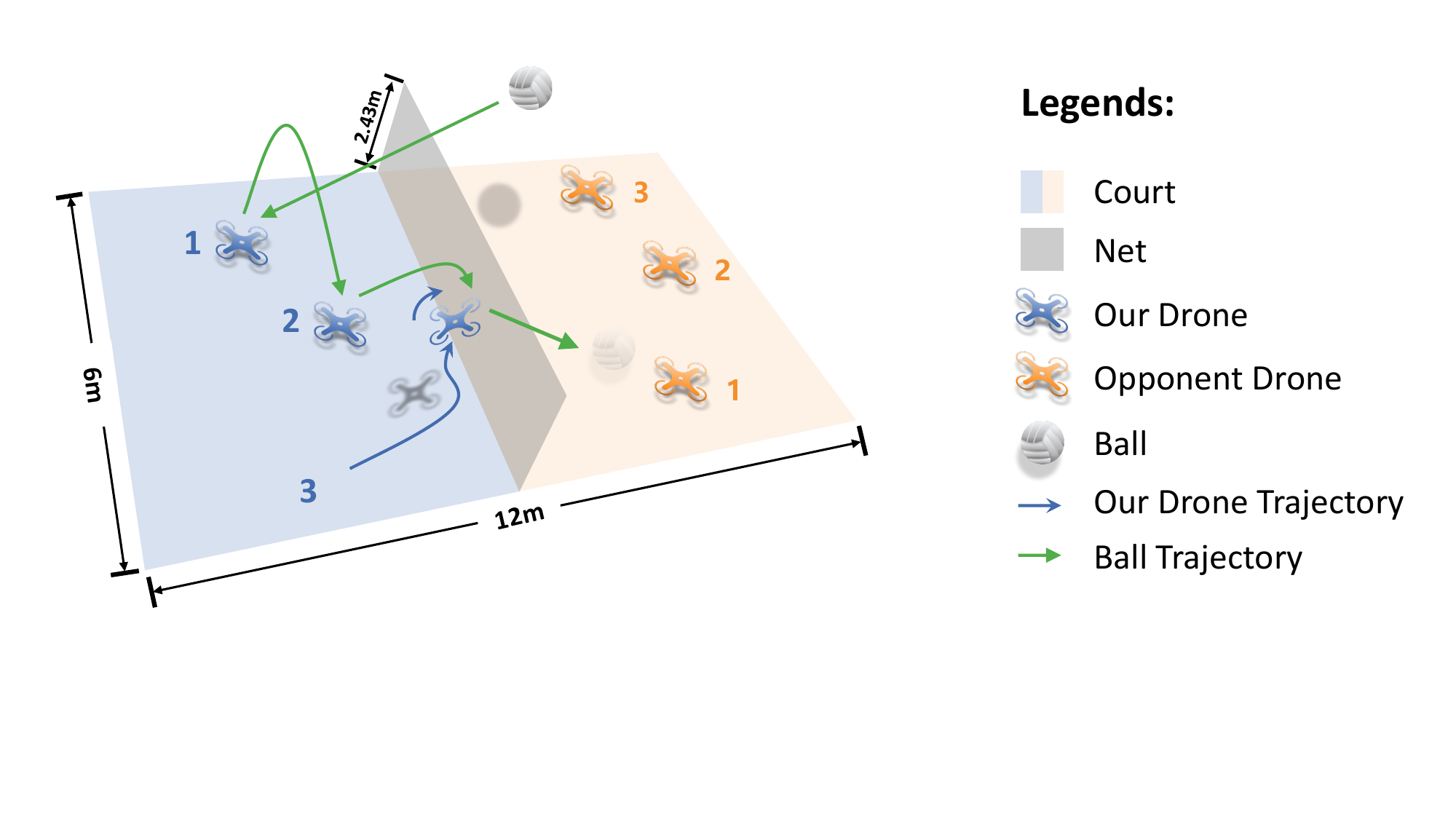}
    \caption{Task overview.}
    \label{fig:task}
  \end{subfigure}%
  \hfill
  \begin{subfigure}[b]{0.45\linewidth}
    \centering
    \includegraphics[width=\linewidth]{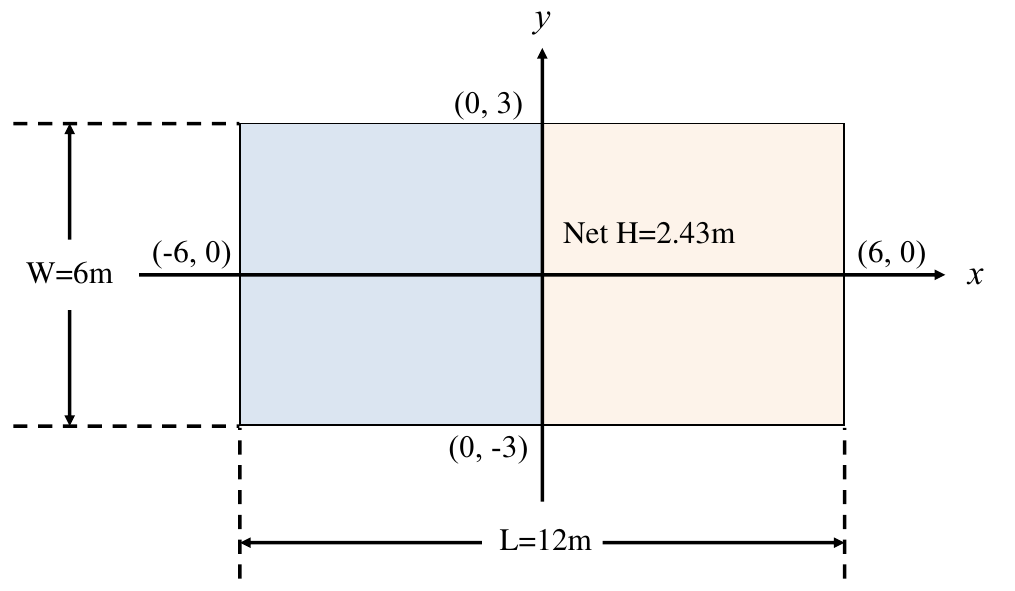}
    \caption{Top-down view of the court layout.}
    \label{fig:layout}
  \end{subfigure}
  \caption{Illustrations of the 3v3 multi-drone volleyball task.}
  \label{fig:appendix/task}
\end{figure}

\subsection{Court}
The 3v3 multi‑drone volleyball task is illustrated in Fig.~\ref{fig:task}, with a top‑down, 2D court layout shown in Fig.~\ref{fig:layout}. The origin lies at the center of the court, and the y‑axis divides it into two equal halves (each team’s side). The x‑axis runs the 12m length (from $x = -6$ to $x = 6$), while the court spans 6m in width along the y‑axis (from $y = -3$ to $y = 3$). A net of height 2.43m stretches horizontally along the line $x=0$, from $(0,-3)$ to $(0,3)$.

\subsection{Drone}
Our primary drone platform is the \textit{Iris} quadrotor ~\cite{furrer2016rotors}, virtually equipped with a 0.2 m radius racket (coefficient of restitution 0.8) for ball striking. The drone's root state used by the simulation framework is described by a 23-dimensional vector containing its position, rotation (quaternion), linear and angular velocities, forward and upward orientation vectors (corresponding to the first two columns of the rotation matrix derived from the quaternion), and normalized rotor speeds. The control dynamics governing the drone's motion arise from its physical configuration and the interplay of forces and torques, as described below:

\begin{align}
\bm{\dot{x}}_W = \bm{v}_W, \quad \bm{\dot{v}}_W = \bm{R}_{WB}f + \bm{g} + \bm{F} \\
\bm{\dot{q}} = \frac{1}{2} \bm{q} \otimes \bm{\omega}, \quad \bm{\dot{\omega}} = \bm{J}^{-1} (\bm{\eta} - \bm{\omega} \times \bm{J} \bm{\omega})
\end{align}

where $\bm{x}_W$ and $\bm{v}_W$ are the position and velocity of the drone in the world frame, $\bm{R}_{WB}$ is the rotation matrix from the body frame to the world frame, $\bm{J}$ is the diagonal inertia matrix, $\bm{g}$ represents gravity, $\bm{q}$ is attitude in quaternion, and $\bm{\omega}$ is the angular velocity. The operator for quaternion multiplication is denoted by $\otimes$. External forces $\bm{F}$ includes aerodynamic drag and downwash effects. The collective thrust $\bm{f}$ and torque $\bm{\eta}$ are calculated based on the thrust of each rotor $\bm{f}_i$ as:

\begin{align}
\bm{f} &= \sum_{i} \bm{R}^{(i)}_B \bm{f}_i \\
\bm{\eta} &= \sum_{i} \bm{T}^{(i)}_B \times \bm{f}_i + sign_i k_i \bm{f}_i
\end{align}

where $\bm{R}^{(i)}_B$ and $\bm{T}^{(i)}_B$ are the orientation and position of the $i$-th rotor in the body frame, $k_i$ is the force-to-moment ratio and $sign_i$ is 1 for clockwise propellers and -1 for counterclockwise propellers.

\subsection{Game Rules}

The multi‑drone game is governed by a set of well‑defined rules that closely mimic the cooperative and competitive dynamics of real‑world volleyball, subject to the following simplifications. Unlike standard volleyball, there is no positional rotation or fixed court zones, so drones do not change roles between rallies, and the three‑meter attack line is omitted.
At the start of each episode, a serving side is randomly selected, and the ball is released in free fall from a point 5m above the midpoint of that team’s baseline, with slight random positional noise.

During the rally phase, teams alternate in attempting to return the ball over the net while obeying two key constraints: (i) each team is allowed no more than three successive hits before the ball must cross to the opponent’s side, (ii) the same drone is not permitted to hit the ball twice in succession. 

The game outcome is determined by a series of terminal conditions that reflect violations of basic physical or tactical rules. A team loses a rally if (i) any of its drones hits the ground (i.e., descends below a predefined altitude threshold), (ii) any drone illegally crosses the net, (iii) any drone performs a racket hit in violation of timing or rule constraints, (iv) the ball lands on their own side of the court, (v) the team hit the ball outside the boundary of the court, or (vi) the ball collides with the net due to their last action. 

In rare cases, both sides violate the rules of the game at the same time. If drones of both sides hit the ground or both sides cross the net in the same simulation step, a draw will be declared. This ensures that the environment maintains fairness while accounting for simultaneity in physical interactions.

\section{Details of Stage I: Low-Level Skill Acquisition}
\label{appendix:2_stage1}

Ensuring smooth transitions between temporally adjacent low‑level primitives—especially when control must switch between different drones or skill types—poses a significant challenge. To address this, we employ policy chaining, iteratively training each primitive so it executes seamlessly under the event‑driven high‑level strategy.
Concretely, we implement seven fundamental low-level skills (\textit{serve, receive, set, attack, pass, hover, goto}) as single-agent reinforcement learning (RL) tasks, each of which is trained separately using Proximal Policy Optimization (PPO)~\cite{schulman2017proximal}. 

Moreover, for ball-interaction skills (\textit{serve, receive, set, attack, pass}), we design a three-stage reward: (i) The before hit phase: optimal preparation incentives through alignment rewards and posture optimization; (ii) The after hit phase: contact quality evaluation based on ball dynamics metrics; (iii) The end phase: bonuses for task success or penalties for incorrect end states. 

\subsection{Low-Level Skills}

\subsubsection{Serve}

\textbf{Description.} 
The \textit{serve} skill is executed by the serving drone of the randomly selected team at the start of each episode. Given a designated target point on the opponent’s court, the drone must hit the ball once to send it close to that location. 
The skill’s duration spans from the beginning of the episode to the moment the serving drone makes contact with the ball.

\textbf{Observation and reward.}
The drone's observation is a vector of dimension $37$ including the drone's root state, the ball's position, the ball's relative position to the drone, the ball's linear velocity, a one‑hot flag indicating whether the drone is permitted to hit the ball, the ball's relative position to the target ball position. The detailed specification of the task's reward function per time step can be found in Table~\ref{tab:app/serve_reward}.

\begin{table}[t]
\centering
    \caption{Reward of the \textit{serve} skill}
    \vspace{1mm}
    \resizebox{\textwidth}{!}{
        \begin{tabular}{ccccc}
            \toprule
            Type & Name & Sparse & Value Range & Description \\
            \midrule
            \multirow{6}{*}{\begin{tabular}[c]{@{}c@{}}Before\\Hit \end{tabular}} 
            & dist\_to\_ball & \xmark & $[0, 0.16]$ &  drone's distance to the ball \\
            & drone\_hit\_ball & \cmark & $\{0, 10\}$ & drone hitting the ball \\
            & penalty\_pos\_x & \cmark & $\{-150, 0\}$ & being too close to the net \\
            & penalty\_pos\_z & \xmark & $[-\infty,0]$ & being too high or too low \\
            & penalty\_roll & \xmark & $[-\infty,0]$ & excessive roll angle\\
            & penalty\_yaw & \xmark & $[-\infty,0]$ & excessive yaw angle\\
            \midrule
            \multirow{1}{*}{\begin{tabular}[c]{@{}c@{}}After Hit\end{tabular}} 
            & dist\_to\_anchor & \xmark & $[0, 16]$ & ball's distance to anchor \\
            \midrule
            \multirow{4}{*}{\begin{tabular}[c]{@{}c@{}}End\\Reward\end{tabular}} 
            & in\_side & \cmark & $\{0, 10\}$ & ball landing in opponent's court \\
            & highest\_ball\_pos & \cmark & $\{0, 1.5\}$ & ball’s peak height exceeding 3m \\
            & penalty\_ground\_collision & \cmark & $\{-0.1, 0\}$ & drone colliding with the ground \\
            & penalty\_wrong\_hit & \cmark & $\{-10, 0\}$ & drone not using racket to hit the ball \\
            \bottomrule
        \end{tabular}
    }
    \label{tab:app/serve_reward}
\end{table}

\subsubsection{Receive}

\textbf{Description.}
The \textit{receive} skill is used by the receiving drone to counter the opponent's serve. During the \textit{receive} skill training, the opponent’s serving drone executes the \textit{serve} skill as part of the environment. The receiving drone needs to hit the ball to the passing drone positioned at the front-left side of its own court.
The skill’s duration spans from the moment the opponent’s serving drone hits the ball to the moment the receiving drone makes contact with it.

\textbf{Observation and reward.}
The drone's observation is a vector of dimension $34$ including the drone's root state, the ball's position, the ball's relative position to the drone, the ball's linear velocity, a one‑hot flag indicating whether the drone is permitted to hit the ball. The detailed specification of the task's reward function per time step can be found in Table~\ref{tab:app/receive_reward}.

\begin{table}[t]
\centering
    \caption{Reward of the \textit{receive} skill}
    \vspace{1mm}
    \resizebox{\textwidth}{!}{
        \begin{tabular}{ccccc}
            \toprule
            Type & Name & Sparse & Value Range & Description \\
            \midrule
            \multirow{6}{*}{\begin{tabular}[c]{@{}c@{}}Before\\Hit \end{tabular}} 
            & dist\_to\_ball & \xmark & $[0, 10]$ &  drone's distance to the ball \\
            & drone\_hit\_ball & \cmark & $\{0, 150\}$ & drone hitting the ball \\
            & penalty\_pos\_x & \cmark & $\{-5, 0\}$ & being too close to the net \\
            & penalty\_pos\_z & \xmark & $[-\infty,0]$ & being too high or too low \\
            & penalty\_roll & \xmark & $[-\infty,0]$ & excessive roll angle\\
            & penalty\_yaw & \xmark & $[-\infty,0]$ & excessive yaw angle\\
            \midrule
            \multirow{3}{*}{\begin{tabular}[c]{@{}c@{}}After\\Hit\end{tabular}} 
            & dist\_to\_anchor & \xmark & $[0, 50]$ & ball's distance to anchor \\
            & ball\_vel\_direction & \xmark & $[0, 150]$ & velocity direction to the anchor \\
            & secondary\_dist\_to\_anchor & \xmark & $[0, 200]$ & height-limited ball's distance to anchor \\
            \midrule
            \multirow{4}{*}{\begin{tabular}[c]{@{}c@{}}End\\Reward\end{tabular}} 
            & in\_side & \cmark & $\{0, 10\}$ & ball landing in opponent's court \\
            & highest\_ball\_pos & \xmark & $[0, 200]$ & ball’s peak height exceeding 3m \\
            & penalty\_ground\_collision & \cmark & $\{-100, 0\}$ & drone colliding with the ground \\
            & penalty\_wrong\_hit & \cmark & $\{-10, 0\}$ & drone not using racket to hit the ball \\
            \bottomrule
        \end{tabular}
    }
    \label{tab:app/receive_reward}
\end{table}

\subsubsection{Set}
\textbf{Description.}
The \textit{set} skill is executed by the setting drone and constitutes the second touch during a team’s turn. During the \textit{set} skill training, the ball is initialized using a Gaussian distribution to simulate the moment it is passed by the teammate. The setting drone is required to redirect the ball to the front-right side of its own court, enabling the attacking drone to perform a strike. The skill’s duration spans from the moment the passing drone contacts the ball to the moment the setting drone hits it.

\textbf{Observation and reward.}
The drone's observation is a vector of dimension $34$ including the drone's root state, the ball's position, the ball's relative position to the drone, the ball's linear velocity, a one‑hot flag indicating whether the drone is permitted to hit the ball. The detailed specification of the task's reward function per time step can be found in Table~\ref{tab:app/set_reward}.

\begin{table}[t]
\centering
    \caption{Reward of the \textit{set} skill}
    \vspace{1mm}
    \resizebox{\textwidth}{!}{
        \begin{tabular}{ccccc}
            \toprule
            Type & Name & Sparse & Value Range & Description \\
            \midrule
            \multirow{5}{*}{\begin{tabular}[c]{@{}c@{}}Before\\Hit \end{tabular}} 
            & dist\_to\_ball & \xmark & $[0, 2]$ &  drone's distance to the ball \\
            & drone\_hit\_ball & \cmark & $\{0, 40\}$ & drone hitting the ball \\
            & penalty\_pos\_xy & \xmark & $[-\infty, 0]$ & being too far from the ideal point \\
            & penalty\_pos\_z & \xmark & $[-\infty,0]$ & being too high or too low \\
            & penalty\_yaw & \xmark & $[-\infty,0]$ & excessive yaw angle\\
            \midrule
            \multirow{2}{*}{\begin{tabular}[c]{@{}c@{}}After\\Hit\end{tabular}} 
            & dist\_to\_anchor & \xmark & $[0, 10]$ & ball's distance to anchor \\
            & drone\_dist\_to\_anchor & \xmark & $[0, 200]$ & drone's distance to anchor \\
            \midrule
            \multirow{3}{*}{\begin{tabular}[c]{@{}c@{}}End\\Reward\end{tabular}} 
            & highest\_ball\_pos & \xmark & $[0, 400]$ & ball’s peak height exceeding 3m \\
            & penalty\_ground\_collision & \cmark & $\{-10, 0\}$ & drone colliding with the ground \\
            & penalty\_wrong\_hit & \cmark & $\{-10, 0\}$ & drone not using racket to hit the ball \\
            \bottomrule
        \end{tabular}
    }

    \label{tab:app/set_reward}
\end{table}

\subsubsection{Attack}
\textbf{Description.}
The \textit{attack} skill is executed by the attacking drone and constitutes the third touch in a team’s sequence. During the \textit{attack} skill training, the setting drone performs the \textit{set} skill as part of the environment. The attacking drone must deliver an effective offensive strike toward one of two predefined target positions on the opponent’s court (left or right). The duration of the skill spans from the moment the setting drone contacts the ball to the moment the attacking drone strikes it.

\textbf{Observation and reward.}
The drone's observation is a vector of dimension $36$ including the drone's root state, the ball's position, the ball's relative position to the drone, the ball's linear velocity, a one‑hot flag indicating whether the drone is permitted to hit the ball, a one‑hot flag indicating the attack direction. The detailed specification of the task's reward function per time step can be found in Table~\ref{tab:app/attack_reward}.

\begin{table}[t]
\centering
    \caption{Reward of the \textit{attack} skill}
    \vspace{1mm}
    \resizebox{\textwidth}{!}{
        \begin{tabular}{ccccc}
            \toprule
            Type & Name & Sparse & Value Range & Description \\
            \midrule
            \multirow{5}{*}{\begin{tabular}[c]{@{}c@{}}Before\\Hit \end{tabular}} 
            & dist\_to\_ball & \xmark & $[0, 1]$ &  drone's distance to the ball \\
            & drone\_hit\_ball & \cmark & $\{0, 20\}$ & drone hitting the ball \\
            & drone\_pos\_z & \xmark & $[0, 400]$ & height of drone hitting the ball  \\
            & penalty\_yaw & \xmark & $[-2\pi,0]$ & excessive yaw angle\\
            & penalty\_roll & \xmark & $[-2\pi,0]$ & excessive roll angle\\
            \midrule
            \multirow{3}{*}{\begin{tabular}[c]{@{}c@{}}After\\Hit\end{tabular}} 
            & dist\_to\_anchor & \xmark & $[0, 150]$ & ball's distance to anchor \\
            & ball\_vel\_x & \cmark & $\{0, 1\}$ & velocity direction of the ball \\
            & ball\_vel\_z & \cmark & $\{0, 150\}$ & downward velocity of the ball \\
            \midrule
            \multirow{6}{*}{\begin{tabular}[c]{@{}c@{}}End\\Reward\end{tabular}} 
            & ball\_hit\_ground & \cmark & $\{0, 5\}$ & ball hit the ground successfully\\
            & ball\_final\_vel & \xmark & $[0, \infty]$ & ball’s velocity hitting the ground \\
            & in\_side & \cmark & $\{0, 10\}$ & ball landing in opponent's court \\
            & penalty\_ground\_collision & \cmark & $\{-10, 0\}$ & drone colliding with the ground \\
            & penalty\_wrong\_hit & \cmark & $\{-10, 0\}$ & drone not using racket to hit the ball \\
            & penalty\_hit\_net & \cmark & $\{-10, 0\}$ & ball hit the net \\
            \bottomrule
        \end{tabular}
    }

    \label{tab:app/attack_reward}
\end{table}

\subsubsection{Pass}
\textbf{Description.}
The \textit{pass} skill is used to counter the opponent's attack and constitutes the first touch in a team’s sequence. 
During the \textit{pass} skill training, the opponent side includes a setting drone that performs the \textit{set} skill and an attacking drone that performs the \textit{attack} skill as part of the environment. 
The passing drone needs to hit the ball to the teammate positioned at the front-left side of its own court.
The duration of the skill spans from the moment the opponent’s attacking drone spikes the ball to the moment the passing drone makes contact with it.

\textbf{Observation and reward.}
The drone's observation is a vector of dimension $36$ including the drone's root state, the ball's position, the ball's relative position to the drone, the ball's linear velocity, a one‑hot flag indicating whether the drone is permitted to hit the ball, a one‑hot flag indicating the opponent's attacking direction. The detailed specification of the task's reward function per time step can be found in Table~\ref{tab:app/pass_reward}.

\begin{table}[t]
\centering
    \caption{Reward of the \textit{pass} skill}
    \vspace{1mm}
    \resizebox{\textwidth}{!}{
        \begin{tabular}{ccccc}
            \toprule
            Type & Name & Sparse & Value Range & Description \\
            \midrule
            \multirow{7}{*}{\begin{tabular}[c]{@{}c@{}}Before\\Hit \end{tabular}} 
            & dist\_to\_ball & \xmark & $[0, 10]$ &  drone's distance to the ball \\
            & drone\_hit\_ball & \cmark & $\{0, 150\}$ & drone hitting the ball \\
            & drone\_vel\_z & \xmark & $[0, 2.5]$ & drone's rising speed \\
            & penalty\_pos\_x & \cmark & $\{-5, 0\}$ & being too close to the net \\
            & penalty\_pos\_z & \xmark & $[-\infty,0]$ & being too high or too low \\
            & penalty\_roll & \xmark & $[-0.5\pi,0]$ & excessive roll angle\\
            & penalty\_yaw & \xmark & $[-0.5\pi,0]$ & excessive yaw angle\\
            \midrule
            \multirow{3}{*}{\begin{tabular}[c]{@{}c@{}}After\\Hit\end{tabular}} 
            & dist\_to\_anchor & \xmark & $[0, 10]$ & ball's distance to anchor \\
            & ball\_vel\_direction & \xmark & $[0, 150]$ & velocity direction to the anchor \\
            & secondary\_dist\_to\_anchor & \xmark & $[0, 200]$ & height-limited ball's distance to anchor \\
            \midrule
            \multirow{4}{*}{\begin{tabular}[c]{@{}c@{}}End\\Reward\end{tabular}} 
            & highest\_ball\_pos & \xmark & $[0, 200]$ & ball’s peak height exceeding 3m \\
            & penalty\_ground\_collision & \cmark & $\{-100, 0\}$ & drone colliding with the ground \\
            & penalty\_not\_hit\_ball & \cmark & $\{-10, 0\}$ & drone not hitting the ball \\
            & penalty\_wrong\_hit & \cmark & $\{-10, 0\}$ & drone not using racket to hit the ball \\
            \bottomrule
        \end{tabular}
    }

    \label{tab:app/pass_reward}
\end{table}

\subsubsection{Hover}

\textbf{Description.}
The \textit{hover} skill is executed immediately after the drone hits the ball. The drone stabilizes at a predetermined altitude and position to prepare for the next move. 
Given that different ball-interaction skills terminate in distinct end states, we define five corresponding hover skills (\textit{serve\_hover, receive\_hover, set\_hover, attack\_hover, pass\_hover}) and train each independently to stabilize the agent after the associated interaction with the same hover reward.
Here, we specifically illustrate the \textit{hover} skill following the \textit{serve} skill. During training, the serving drone performs the \textit{serve} skill as part of the environment, and immediately after it contacts the ball, its policy switches to the \textit{hover} skill.

\textbf{Observation and reward.}
The drone's observation is a vector of dimension $26$ including the drone's relative position to the target point, the drone's root state except position, the drone's relative orientation to the ideal state. The detailed specification of the task's reward function per time step can be found in Table~\ref{tab:app/hover_reward}.

\begin{table}[t]
\centering
    \caption{Reward of the \textit{hover} skill}
    \vspace{1mm}
    \resizebox{\textwidth}{!}{
        \begin{tabular}{ccccc}
            \toprule
            Type & Name & Sparse & Value Range & Description \\
            \midrule
            \multirow{3}{*}{\begin{tabular}[c]{@{}c@{}}Hover\\Reward\end{tabular}} 
            & drone\_pos & \xmark & $[0, 3]$ & drone's relative position to the target point\\
            & drone\_up\ & \xmark & $[0, 3]$ & drone upright position \\
            & drone\_spin & \xmark & $[0, 3]$ & drone spin suppression \\
            \bottomrule
        \end{tabular}
    }

    \label{tab:app/hover_reward}
\end{table}

\subsubsection{GoTo}
\textbf{Description.}
The \textit{goto} skill is performed after the drone has stabilized and involves navigating to a designated target position at a fixed altitude, preparing the drone for an upcoming hit. In training, the drone’s initial and target positions are randomized to encourage generalization across various scenarios.

\textbf{Observation and reward.}
The drone's observation is a vector of dimension $26$ including the drone's relative position to the target point, the drone's root state except position, the drone's relative orientation to the ideal state. The detailed specification of the task's reward function per time step can be found in Table~\ref{tab:app/goto_reward}.

\begin{table}[t]
\centering
    \caption{Reward of the \textit{goto} skill}
    \vspace{1mm}
    \resizebox{\textwidth}{!}{
        \begin{tabular}{ccccc}
            \toprule
            Type & Name & Sparse & Value Range & Description \\
            \midrule
            \multirow{5}{*}{\begin{tabular}[c]{@{}c@{}}Hover\\Reward\end{tabular}} 
            & drone\_pos & \xmark & $[0, 1]$ & drone's relative position to the target point\\
            & drone\_up\ & \xmark & $[0, 1]$ & drone upright position \\
            & drone\_spin & \xmark & $[0, 1]$ & drone spin suppression \\
            & drone\_effort & \xmark & $[0, 0.1]$ & hover with low effort \\
            \bottomrule
        \end{tabular}
    }

    \label{tab:app/goto_reward}
\end{table}

\subsection{Training Result}

\begin{figure}[ht]
    \centering
    \includegraphics[width=1\linewidth]{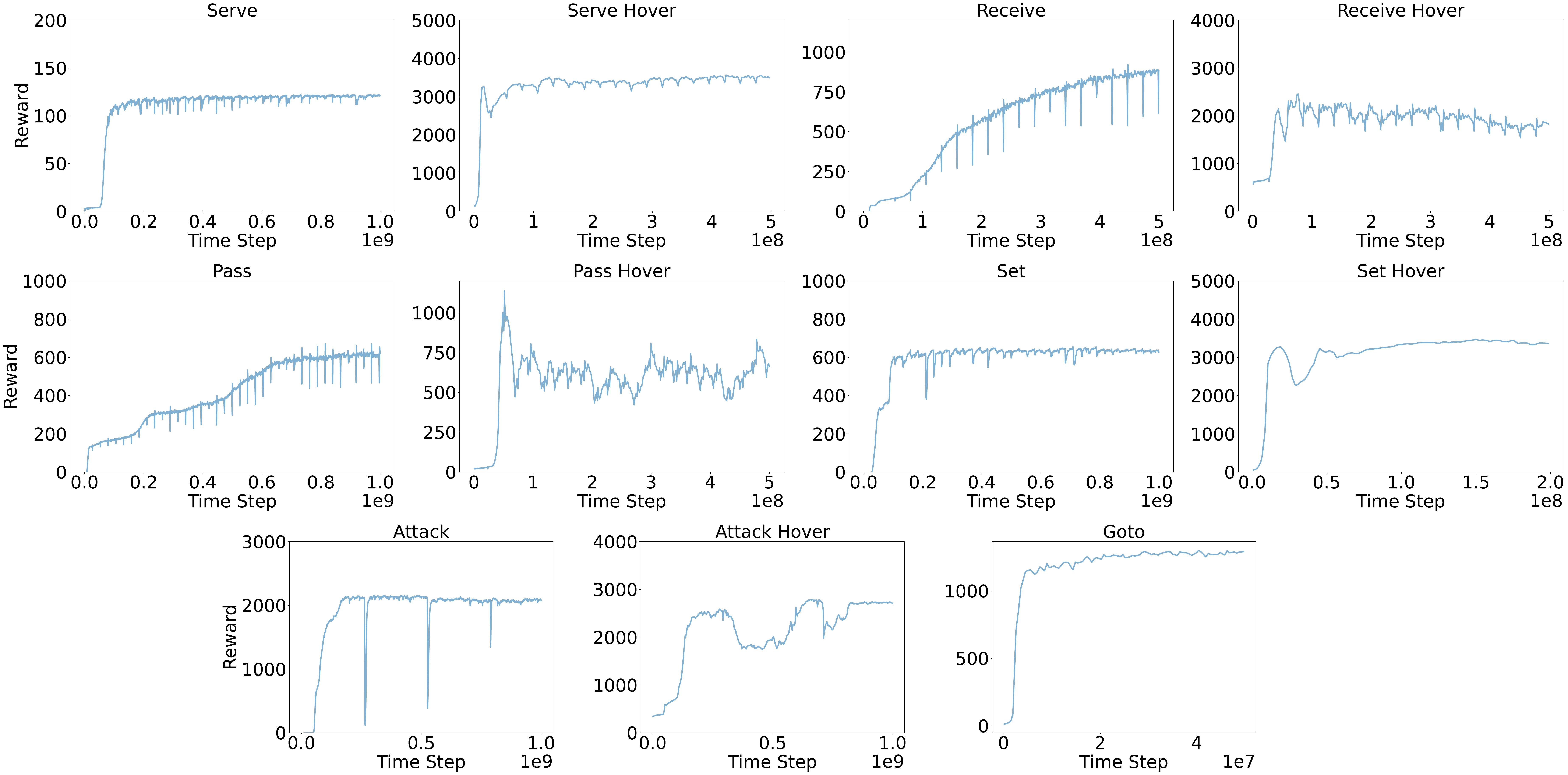}
    \caption{The training results of low-level skills.}
    \label{fig:low_level_reward}
\end{figure}

The training results of low-level skills are shown in Fig.~\ref{fig:low_level_reward}.

\subsection{Case Study}
\begin{figure}[t]
  \centering
  \begin{subfigure}[b]{0.16\textwidth}
    \centering
    \includegraphics[width=\linewidth]{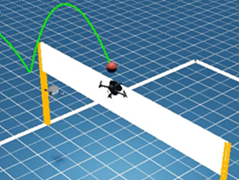}
    \caption{}
  \end{subfigure}\hfill
  \begin{subfigure}[b]{0.16\textwidth}
    \centering
    \includegraphics[width=\linewidth]{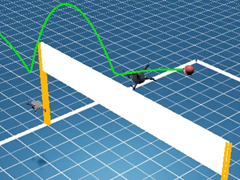}
    \caption{}
  \end{subfigure}\hfill
  \begin{subfigure}[b]{0.16\textwidth}
    \centering
    \includegraphics[width=\linewidth]{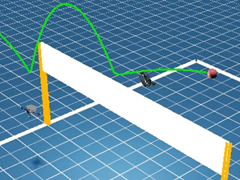}
    \caption{}
  \end{subfigure}\hfill
  \begin{subfigure}[b]{0.16\textwidth}
    \centering
    \includegraphics[width=\linewidth]{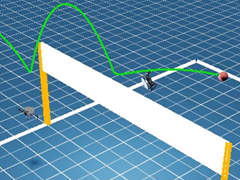}
    \caption{}
  \end{subfigure}\hfill
  \begin{subfigure}[b]{0.16\textwidth}
    \centering
    \includegraphics[width=\linewidth]{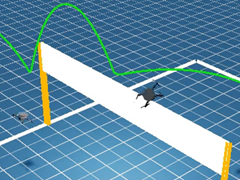}
    \caption{}
  \end{subfigure}\hfill
  \begin{subfigure}[b]{0.16\textwidth}
    \centering
    \includegraphics[width=\linewidth]{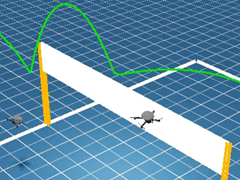}
    \caption{}
  \end{subfigure}
  \caption{Sequence of six frames, sampled sequentially in time from the start to the completion of the front‑flip attack.}
  \label{fig:attack_flip_hit}
\end{figure}

When training the \textit{attack} and \textit{attack\_hover} skills through policy chaining, we observed the emergence of a front‑flip attack (Fig.~\ref{fig:attack_flip_hit}). 
Executing a front flip is particularly challenging for a drone, since it requires precise timing, rapid attitude changes, and tight control over thrust to complete a full somersault without losing stability.
Remarkably, even without explicitly designing for this maneuver, policy chaining can facilitate learning to perform a front flip to attack the ball. 
Moreover, performing a front‑flip attack allows the drone to impart additional downward and forward momentum at the moment of impact, resulting in a higher ball speed.
The finding shows that policy chaining not only facilitates complex skill discovery but also naturally promotes the emergence of aggressive, high‑performance behaviors through the composition of simpler skills.

\subsection{Hyperparameters}
The PPO~\cite{schulman2017proximal} hyperparameters applied to all low‑level skill training tasks are listed in Table~\ref{tab:app/LLS_params}.  All parameters in the table are used unchanged except for the actor and critic learning rates:
\begin{itemize}
\item \textit{receive}, \textit{pass}, and \textit{pass\_hover}: actor learning rate \& critic learning rate = $1\times10^{-5}$
\item \textit{receive\_hover}: actor learning rate \& critic learning rate = $1\times10^{-4}$
\end{itemize}
\begin{table}[t]
\centering
    \caption{Hyperparameters used for PPO in low-level skill training.}
    \vspace{1mm}
    \resizebox{\textwidth}{!}{
        \begin{tabular}{l c | l c | l c}
            \toprule
            hyperparameters & value & hyperparameters & value & hyperparameters & value \\
            \midrule
            optimizer & Adam & max grad norm & $10$ & entropy coef & $0.001$\\
            buffer length & $64$ & num minibatches & $16$ & PPO epochs & $4$ \\
            value norm & ValueNorm1 & clip param & $0.1$ & normalize advantages & True\\
            actor learning rate & $5\times10^{-4}$ & hidden sizes & $[256,128,128]$ & GAE lambda & $0.95$ \\
            critic learning rate & $5\times10^{-4}$ & gain & $0.01$ & GAE gamma & $0.995$ \\
            max episode length & $500$ & num envs & $4096$ & train steps & $1\times10^{9}$ \\
            \bottomrule
        \end{tabular}
    }

    \label{tab:app/LLS_params}
\end{table}

\section{Details of Stage II: High-Level Strategy Pretraining}

The high-level strategy is invoked only at discrete events, namely when the ball is struck or crosses the net, while low-level skills execute continuously between those moments.
In Stage II, we freeze the low-level skills and train the high-level strategy using $\text{PSRO}_\text{Nash}$~\citep{lanctot2017unified}, an iterative game-theoretic self-play method where policies compete against opponents sampled from a Nash equilibrium distribution. 
To address the sparsity of high-level decisions, we employ sample reallocation, focusing training on event-triggered timesteps and redistributing rewards across intervals. 
The high-level strategy is implemented as a centralized multi‑layer perceptron (MLP) with three output heads, each issuing commands to a specific drone, thereby enabling coordinated team-level decision making.
The entire framework is trained using MAPPO~\cite{yu2022surprising}.

\subsection{High-Level Strategy}

\textbf{Observation.}
The high-level strategy's observation is a vector of dimension $100$ including the root states (position, rotation, linear velocity, angular velocity) of 6 drone agents (3 on each side), the ball's position, the ball's linear velocity, binary flags indicating which drone agents have already hit the ball, ball side information, a one-hot representation of the current game phase.

\textbf{Action.}
The high‑level action space is organized as a multi‑head structure with three heads of sizes 6, 3, and 4, corresponding to the passing, setting, and attacking drone roles.  Each head defines an independent categorical distribution: its output logits are converted to normalized action probabilities via a softmax activation. Prior to discretizations, the logits for each head are sampled from Gaussian distributions in a latent policy space, thereby facilitating probabilistic exploration.

\textbf{Reward.}
The reward for the high-level strategy $\pi_j^H$ at timestep $t$ is defined as:
\begin{equation}
    r_{j,t}^H = c_1 \times \mathrm{win\_or\_lose}_j + c_2 \times \mathrm{racket\_hit\_ball}_j,
\end{equation}
where $\mathrm{win\_or\_lose}_j = 1$ if team $j$ wins, $-1$ if it loses, and $0$ otherwise; $\mathrm{racket\_hit\_ball}_j = 1$ if any racket of team $j$ contacts the ball at timestep $t$, and $0$ otherwise.

\subsection{Training Result}
\begin{figure}[ht]
    \centering
    \includegraphics[width=0.4\linewidth]{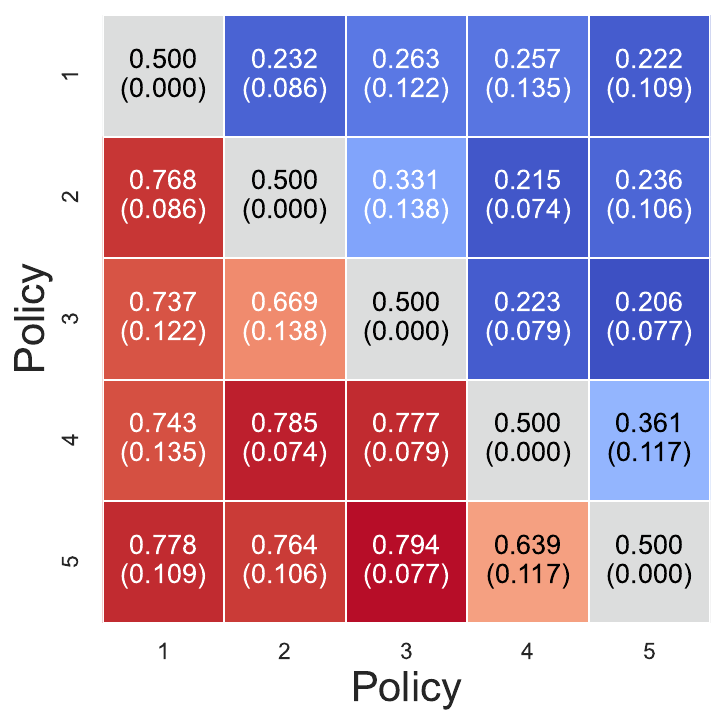}
    \caption{Win-rate heatmap illustrating the evolution of high-level strategy training in Stage II.}
    \label{fig:appendix/stage_2_heatmap}
\end{figure}

The results of high‑level strategy training are shown in Fig.~\ref{fig:appendix/stage_2_heatmap}. In our experiments, we set the maximum population size to five, yielding five iterative training phases; each phase terminates when either a preset number of training steps is reached or a win‑rate threshold is met. The lower triangular portion of the matrix reports the win rate of each row policy against each column policy (i.e. the probability that the row policy prevails). Because the game is symmetric and zero‑sum, the upper‑triangular entries are complementary, such that each pair of win rates sums to one.

\subsection{Hyperparameters}
The hyperparameters used for $\text{PSRO}_\text{Nash}$ in high-level strategy training are shown in Table~\ref{tab:app/HLS_params}.
\begin{table}[ht]
\centering
    \caption{Hyperparameters used for $\text{PSRO}_\text{Nash}$ in high-level strategy training.}
    \vspace{1mm}
    \resizebox{\textwidth}{!}{
        \begin{tabular}{l c | l c | l c}
            \toprule
            hyperparameters & value & hyperparameters & value & hyperparameters & value \\
            \midrule
            optimizer & Adam & max grad norm & $10$ & entropy coef & $0.001$\\
            buffer length & $64$ & num minibatches & $16$ & PPO epochs & $4$ \\
            value norm & ValueNorm1 & clip param & $0.1$ & normalize advantages & True\\
            actor learning rate & $1\times10^{-4}$ & hidden sizes & $[256,128,128]$ & GAE lambda & $0.95$ \\
            critic learning rate & $1\times10^{-4}$ & gain & $0.01$ & GAE gamma & $0.995$ \\
            max episode length & $1500$ & num envs & $512$ & min iteration steps & $1.6384\times10^6$ \\
            max population size & $5$ & win rate threshold & $0.9$ & max iteration steps & $1.6384\times10^8$ \\
            \bottomrule
        \end{tabular}
    }
    \label{tab:app/HLS_params}
\end{table}

\section{Details of Stage III: Co-Self-Play}
Stage III consists of two steps (Fig.~\ref{fig:stage3}): the low-level skill refinement step and the high-level strategy relearning step. 
In the low-level skill refinement step, the high-level strategy together with every low-level skill except \textit{serve} and \textit{receive} are co-optimized, resulting in updated low-level skills, which are then added to the existing low-level skill pool. 
The previous high-level strategy becomes invalid due to the change in the size of the skill pool.
Therefore, the high-level strategy relearning step involves retraining the high-level strategy to accommodate the augmented low-level skill pool. 
To analyze the low-level skill refinement step, we conduct two ablation studies: one on the refinement schedule (Appendix~\ref{appendix:stage3/training_schedule}) and the other on the KL divergence penalty term (Appendix~\ref{appendix:stage3/kl}). For the high-level strategy relearning step, we perform an ablation study on the initialization of high-level strategy parameters (Appendix~\ref{appendix:stage3/param_init}).

\subsection{Ablation Study on Low-Level Skill Refinement Schedule}
\label{appendix:stage3/training_schedule}

\begin{figure}[t]
  \centering
  \begin{subfigure}[b]{0.3\textwidth}
    \centering
    \includegraphics[width=\linewidth]{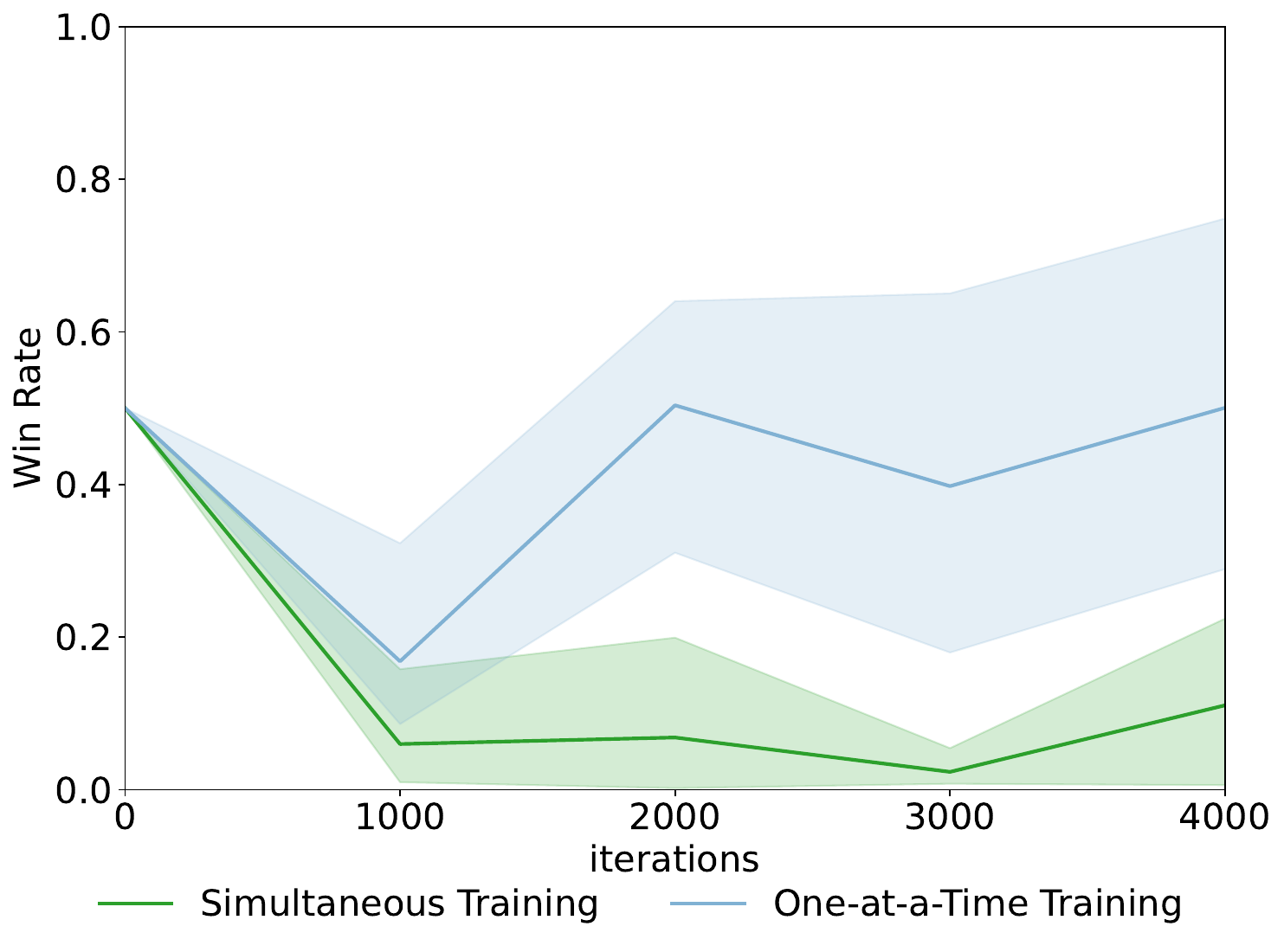}
    \caption{Comparison of win rates under different co-optimization training schedules.}
    \label{fig:appendix/stage3/training_schedule}
  \end{subfigure}\hfill
  \begin{subfigure}[b]{0.3\textwidth}
    \centering
    \includegraphics[width=\linewidth]{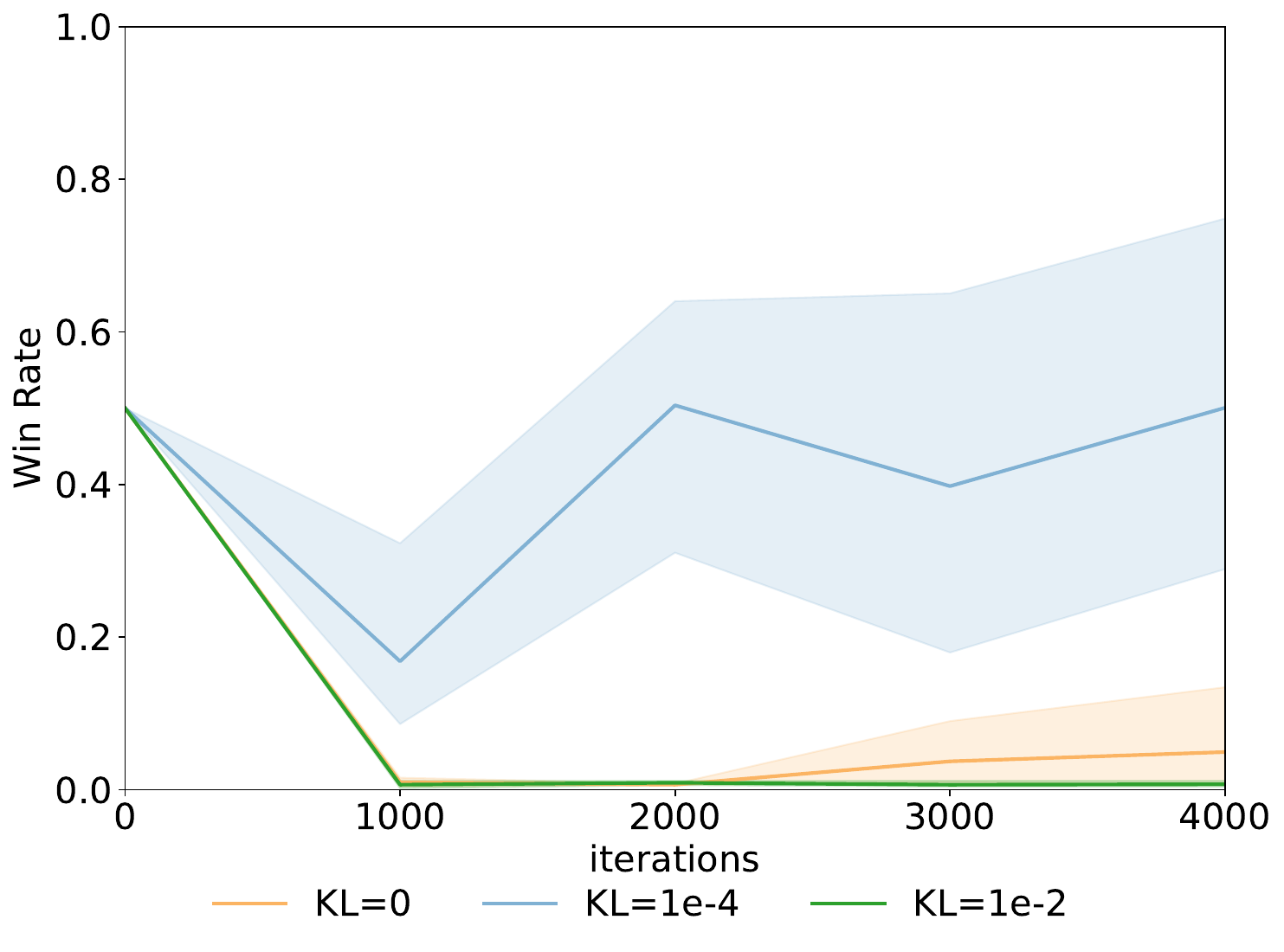}
    \caption{Comparison of win rates under different KL divergence coefficients.}
    \label{fig:appendix/stage3/kl}
  \end{subfigure}\hfill
  \begin{subfigure}[b]{0.34\textwidth}
    \centering
    \includegraphics[width=\linewidth]{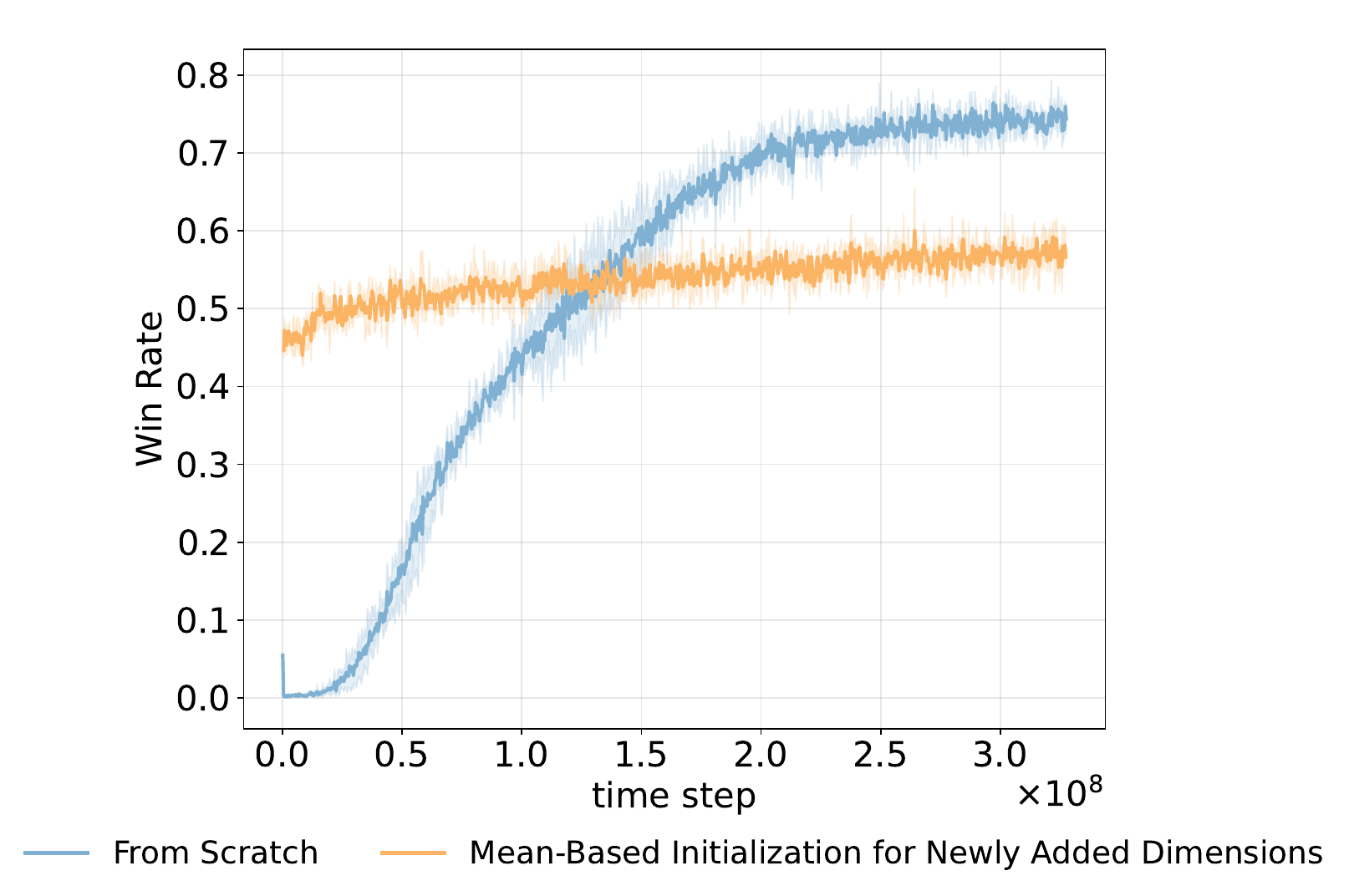}
    \caption{Comparison of win rates under different high-level strategy parameter initializations.}
    \label{fig:appendix/stage3/param_init}
  \end{subfigure}
  \caption{Ablation study results in Stage III.}
\end{figure}

Instead of jointly co-optimizing the high-level strategy with all low-level skills simultaneously, we co-optimize the high-level strategy with one low-level skill at a time (Algo.~\ref{alg:joint_optimization}).
We compare the two co-optimization methods by evaluating the win rates of the saved checkpoints against the high-level strategy and low-level skills obtained after Stage II, under identical settings. These settings include the reward structure, maximum number of iterations ($M = 4000$), checkpoint saving interval ($N = 1000$), and win rate threshold ($t = 0.7$).
Each iteration collects $32768$ sampled steps. All results are averaged over three random seeds.
As shown in Fig.~\ref{fig:appendix/stage3/training_schedule}, jointly co-optimizing the high-level strategy with all low-level skills leads to near-zero win rates. In contrast, co-optimizing the high-level strategy with one low-level skill at a time can yield checkpoints that achieve win rates above 50\%.
Since MARL is inherently non-stationary, simultaneously updating all low-level skills destabilizes the learning process. In comparison, optimizing one low-level skill at a time reduces non-stationarity and enables more stable and effective learning.

\begin{algorithm}[ht]
\caption{High-level strategy and one-at-a-time low-level skill joint optimization.}
\label{alg:joint_optimization}
    \begin{algorithmic}[1]
        \REQUIRE Low-level skill set $\{\pi^L\}$, high-level policy $\pi^H$, number of max iterations $M$, checkpoint saving interval $N$, win rate threshold $t$
        \ENSURE Augmented low-level skill set $\{\pi^L_\text{aug}\}$
    
        \STATE Initialize $\Pi_{\text{new}} \gets \emptyset$
    
        \FOR{each ${\pi^L} \in \{\pi^L\}$}
            \STATE Freeze parameters of all other low-level policies in $\{\pi^L\}$
            \STATE ${\pi^L}' \gets \pi^L $
            \STATE ${\pi^H}' \gets \pi^H $
            \FOR{$i = 1$ to $M$}
                \STATE Jointly optimize ${\pi^L}'$ and ${\pi^H}'$ through self-play against frozen $\{\pi^L\}$ and $\pi^H$
                \IF{$i \bmod N = 0$}
                    \STATE Save a checkpoint of ${\pi^L}'$ and ${\pi^H}'$
                \ENDIF
            \ENDFOR
    
            \STATE Evaluate all saved checkpoints of ${\pi^L}'$ and ${\pi^H}'$ win rates against frozen $\{\pi^L\}$ and $\pi^H$
            \STATE Select the checkpoint of ${\pi^L}'$ with the highest win rate, denoted as $\pi^L_{\text{best}}$
            \IF{win rate of $\pi^L_{\text{best}} > t$}
                \STATE Add $\pi^L_{\text{best}}$ to $\Pi_{\text{new}}$
            \ENDIF
        \ENDFOR
    
        \RETURN $\{\pi^L\} \cup \Pi_{\text{new}}$
    \end{algorithmic}
\end{algorithm}

\subsection{Ablation Study on KL Divergence Penalty}
\label{appendix:stage3/kl}

As stated in Sec.~\ref{sec:stage3}, we incorporate a KL divergence penalty term into the low-level skill reward, in addition to the game-result-related reward shared with the high-level strategy. Here, we conduct a detailed ablation study on the KL divergence penalty to investigate its influence on training performance. 
As shown in Fig.~\ref{fig:appendix/stage3/kl}, and following Algo.~\ref{alg:joint_optimization}, we compare different settings of the KL penalty by evaluating the win rates of the saved checkpoints against the high-level strategy and low-level skills obtained after Stage II.
We observe that a large KL penalty coefficient (e.g., $1\mathrm{e}{-2}$) causes the policy performance to collapse rapidly; conversely, without a KL penalty (coefficient = 0), the policy hovers only slightly above zero win rate. Only with a moderate coefficient (e.g., $1\mathrm{e}{-4}$) does the policy yield checkpoints achieving win rates above 50\%.
This is because, without a KL penalty, the reward signal for low-level strategies becomes sparse, significantly slowing down learning. On the other hand, an overly large KL penalty forces the policy to remain too close to the original one, which leads to accumulated error over time as the policy rigidly imitates previous behaviors without adaptation.


\subsection{Ablation Study on High-Level Strategy Parameter Initialization}
\label{appendix:stage3/param_init}

As the previous high-level strategy becomes invalid due to changes in the size of the low-level skill pool, it is necessary to retrain the high-level strategy to accommodate the augmented skill set.
The initialization of this new high-level strategy is a critical factor influencing the overall performance in Stage III.
We compare two different initialization methods for the high-level strategy. 
The first is a straightforward approach that trains the high-level strategy from scratch. 
In the second approach, we expand each actor head’s parameter vector to match the larger action space by copying over its original values. Any newly added dimensions are then initialized to the mean of that head’s original parameters.
In both settings, we train the new high-level strategy against the frozen high-level strategy and low-level skills obtained after Stage II.
As shown in Fig.~\ref{fig:appendix/stage3/param_init}, although training from scratch requires more time steps to converge, it ultimately achieves a higher win rate. In contrast, the parameter-averaging initialization method tends to converge to a suboptimal local solution.


\section{Baselines}
\label{appendix:5_baselines}

We evaluate five baseline methods for this multi-drone game, including four non-hierarchical game-theoretic approaches and one rule-based hierarchical baseline adapted from VolleyBots~\citep{xu2025volleybots}. The game-theoretic baselines follow estabilished literature~\cite{zhang2024survey} and comprise Self-Play (SP)~\citep{samuel1959some}, Fictitious Self-Play (FSP)~\citep{heinrich2015fictitious}, Policy-Space Response Oracles~\citep{lanctot2017unified} with Nash meta-solver ($\text{PSRO}_{\text{Nash}}$), and PSRO with uniform meta-solver ($\text{PSRO}_{\text{Uniform}}$). Additionally, we incorporate a rule-based hierarchical baseline from VolleyBots.

SP establishes the foundation by training agents against their current policy versions. FSP extends this by maintaining a historical policy pool, forcing agents to adapt to an averaged opponent strategy. $\text{PSRO}_{\text{Nash}}$ employs Nash equilibrium weighting for strategic policy selection, while $\text{PSRO}_{\text{Uniform}}$ simplifies this by sampling policies uniformly at each iteration.

The rule-based hierarchical baseline from VolleyBots implements a structured volleyball-inspired gameplay pipeline. In this system, the serving drone initiates play by executing a serve to a specific target position on the opponent's court. The receiving team then follows a deterministic sequence: (i) the passing drone receives the ball and directs it to the setting drone, (ii) the setting drone positions the ball for optimal attack placement, and (iii) the attacking drone executes an offensive strike.

\section{Evaluation Metrics}
\label{appendix:6_evaluation_metrics}
\subsection{Win Rate}
\label{appendix:win_rate}
The win rate is computed by evaluating two populations across 500 independent episodes. At the beginning of each episode, one strategy is sampled from each population according to a specified sampling distribution. The two strategies then compete, and the outcome (win/loss/draw) is recorded. The final win rate represents the proportion of episodes won by one population relative to the other, averaged over 3 random seeds to ensure statistical reliability.

\subsection{Nash-Averaging Evaluator}

While head-to-head win rates fail to reflect absolute policy strength in non-transitive games, and Elo ratings~\cite{elo1978rating} exhibit sensitivity to the composition of the policy population, Nash-averaging evaluator, introduced by~\citet{liu2019emergent}, provides a robust alternative. 
This method selects a randomly sampled set of policies, then constructs a payoff matrix from pairwise match-ups, and finally computes a Nash equilibrium (NE). The resulting equilibrium weights yield a comparatively population-invariant measure of relative performance, mitigating biases induced by cyclic dominance. So we adopt this as an evaluation metric. The details are as follows.

To compute a Nash-averaging evaluator, we first randomly select 10 different policy checkpoints: 1 each from SP, FSP, $\text{PSRO}_{\text{Uniform}}$, and $\text{PSRO}_{\text{Nash}}$, along with 3 from Stage II and 3 from Stage III of training. All checkpoints are drawn from completely independent training runs with different random seeds. Each policy pair then competes in 500 independent episodes under identical conditions, with the row policy's win rate against the column policy recorded as the corresponding matrix element. All results are averaged across three seeds to ensure statistical reliability. Then, we get a $10\times10$ win rate heatmap (Fig.~\ref{fig:nash_averaging_sub}).

The meta policy is derived by computing the NE of the $10\times10$ win rate matrix. Since the underlying game is zero-sum, win rates (probabilistic outcomes) must first be converted into zero-sum payoff values within the range $[-1, 1]$, where one agent's gain is the other's loss. Specifically, the conversion follows:

\begin{equation}
    \text{payoffs}[i,j] = 2 \times \text{win\_rates}[i,j] - 1
\end{equation}

We then compute each player's NE-based optimal mixed strategy, ensuring neither player can gain a higher expected payoff by unilaterally deviating from their strategy. This computed mixed strategy is the Nash-averaging evaluator we used in Sec.~\ref{sec:quant_analysis_of_stage_three}. We then evaluate the Stage II and Stage III population against this Nash-averaging evaluator in the same way described in Appendix~\ref{appendix:win_rate}.

\section{Preliminary Real-World Experiments}
\label{appendix:7_preliminary_real_world_experiments}

\begin{figure}[t]
  \centering
  \hfill
  \subcaptionbox{Drone setup\label{fig:a}}
    [.31\linewidth]{\includegraphics[width=\linewidth]{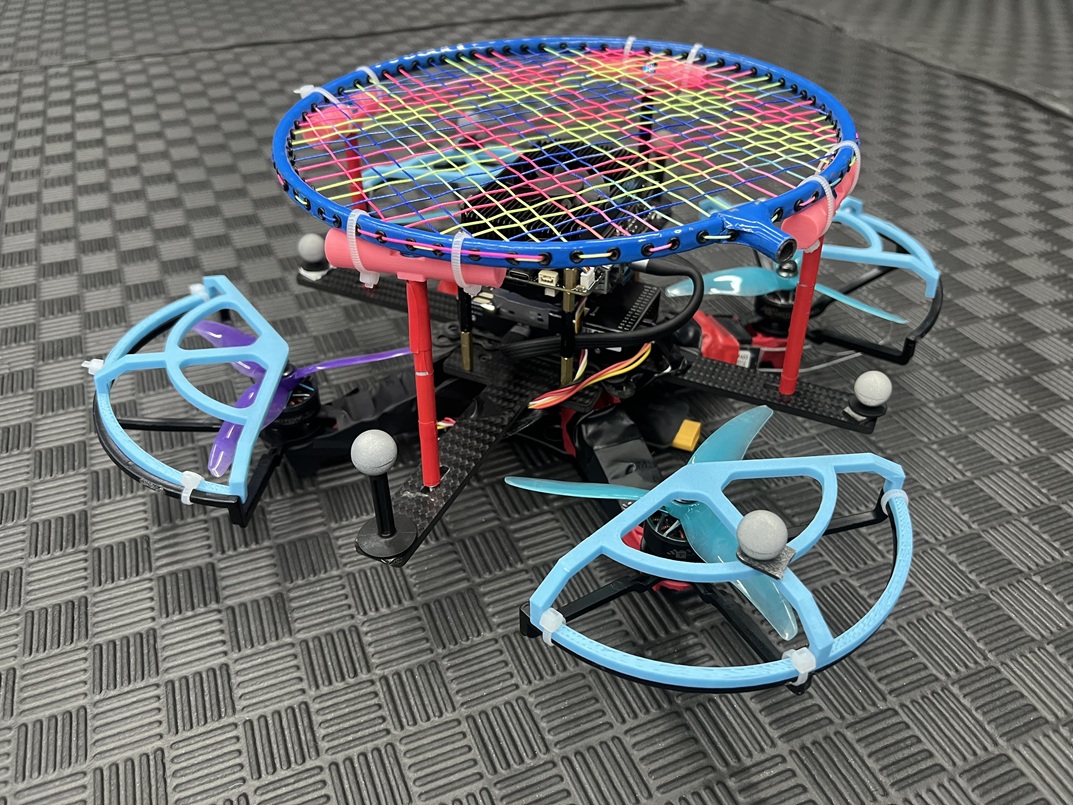}}\hfill
  \subcaptionbox{\textit{Serve}\label{fig:b}}
    [.295\linewidth]{\includegraphics[width=\linewidth]{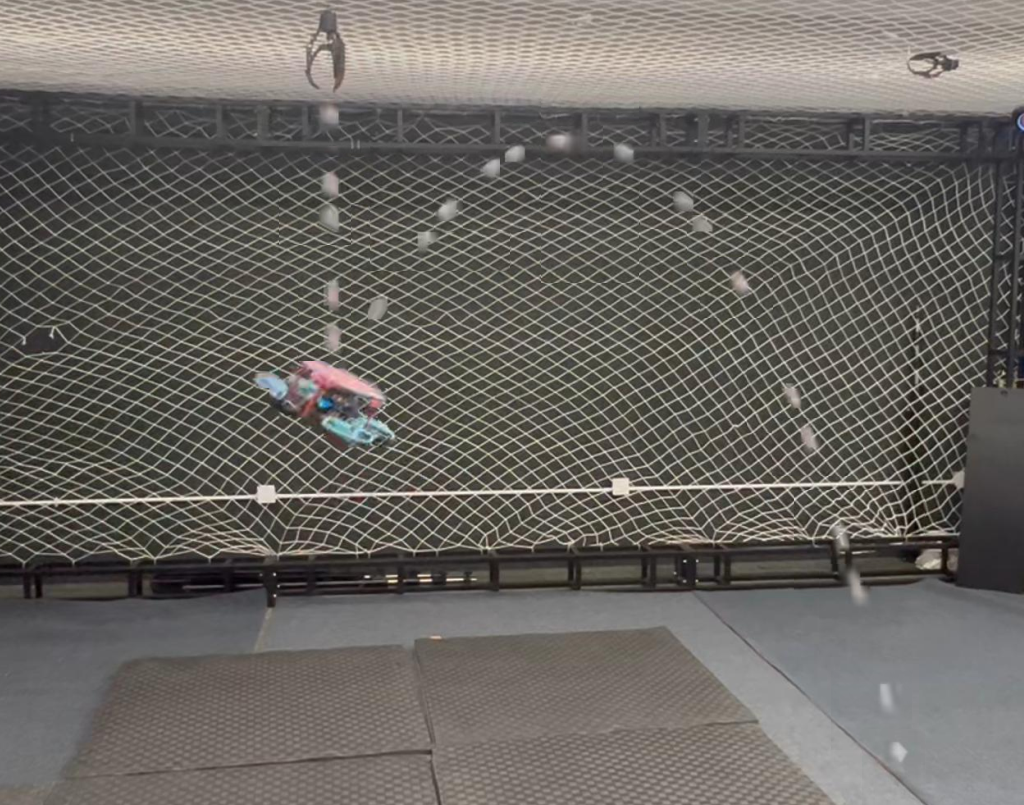}}\hfill
  \subcaptionbox{\textit{Solo Bump}\label{fig:c}}
    [.212\linewidth]{\includegraphics[width=\linewidth]{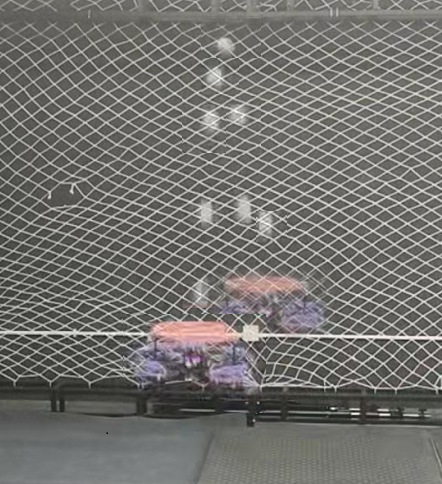}}\hfill\mbox{}

  \hfill
  \subcaptionbox{Ball trajectory in \textit{Serve} \label{fig:d}}
    [.37\linewidth]{\includegraphics[width=\linewidth]{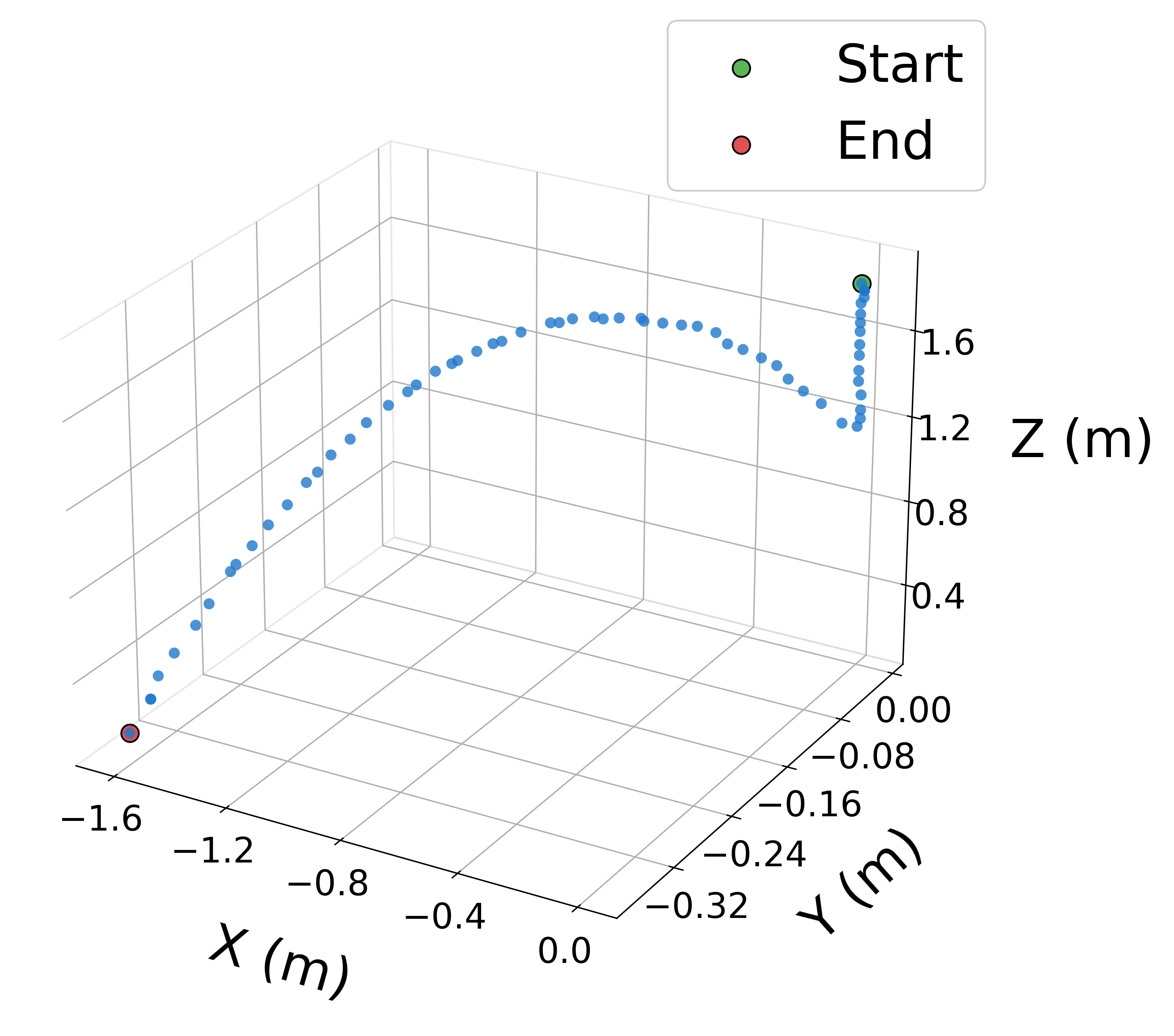}}\hfill
  \subcaptionbox{Height variation in \textit{Solo Bump}\label{fig:e}}
    [.48\linewidth]{\includegraphics[width=\linewidth]{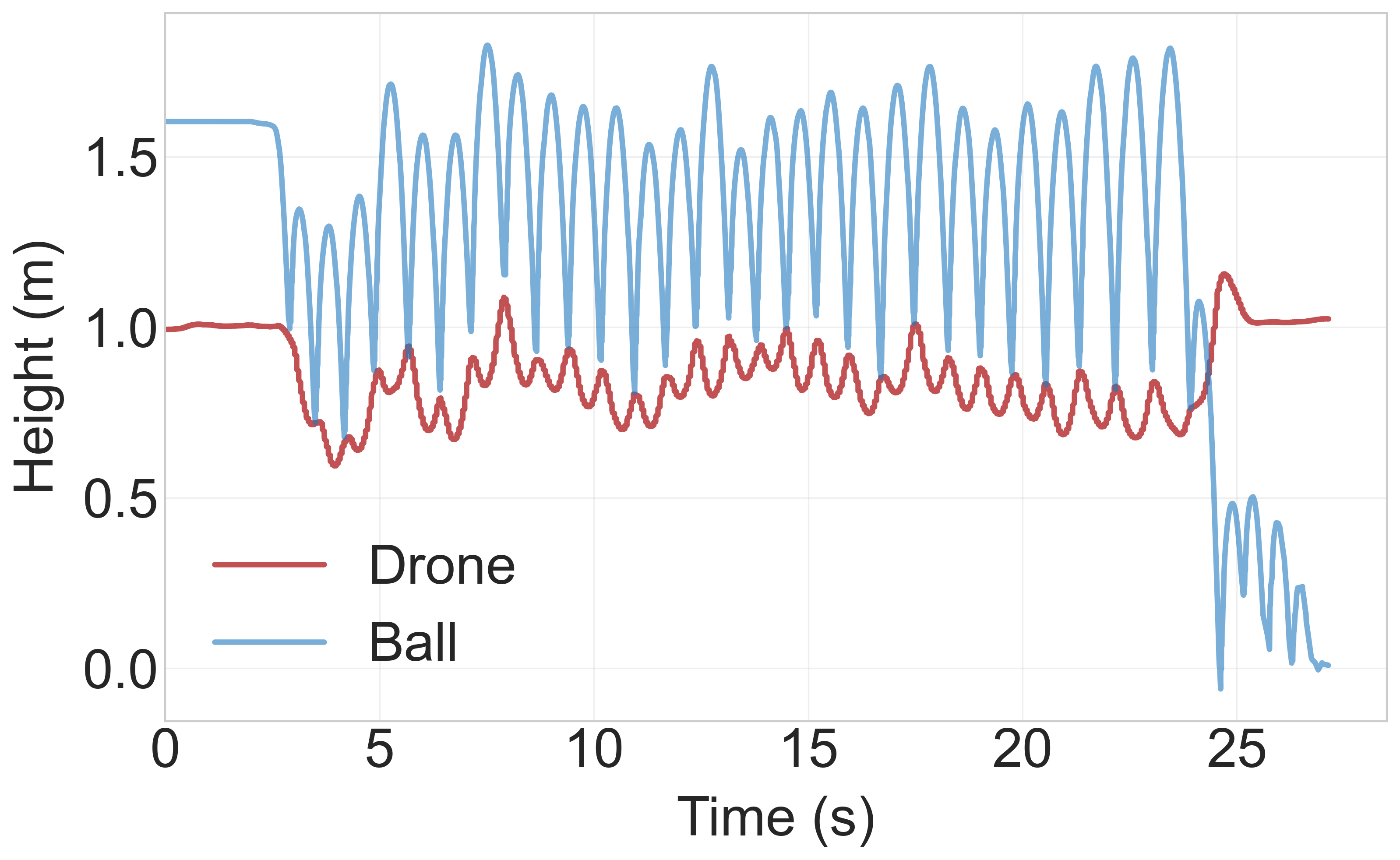}}\hfill\mbox{}
  \caption{Real-world experiments.}
  \label{fig:sim2real}
\end{figure}

Although the current method is not yet sufficient for full 3v3 zero-shot deployment, we incorporate key real-world factors in simulation.
Our preliminary Sim2Real adaptations include:
\begin{itemize}
    \item We support two action space options. In addition to the 50Hz PRT, we also support 50Hz CTBR policy output followed by an 800Hz PID controller.
    \item We add both domain randomization (initial states and restitution coefficient) and observation latency.
\end{itemize}

We further conduct preliminary real-world experiments to demonstrate the zero-shot Sim2Real potential of our approach on learned low-level skills, such as the \textit{Serve} skill and a \textit{Solo Bump} task, in which the drone is required to bump the ball repeatedly to showcase its agility and stability.
As shown in Fig.~\ref{fig:a}, to stay within the quadrotor’s thrust margin, we mount a badminton racket (diameter 22cm, mass 60g) on top of the drone and use a small foam ball (diameter 4cm, mass 47g). 
Both the drone and the ball are tracked by a motion-capture system: the drone, modeled as a rigid body, receives position and orientation directly from mocap, while an Extended Kalman filter fuses these data with PX4 IMU readings to estimate velocity; the ball, treated as a point mass, also gets position from mocap, and a Kalman filter differentiates these measurements to recover velocity.
The policies are deployed on the onboard Nvidia Orin processor.
In zero-shot trials, the low-level skills such as \textit{Serve} are executed successfully (Fig.~\ref{fig:b} and~\ref{fig:d}) and the \textit{Solo Bump} policy achieves 29 consecutive bumps (Fig.~\ref{fig:c}~and~\ref{fig:e}).
These results provide promising evidence for the physical plausibility and transferability of our approach.

\end{document}